\documentclass[lettersize,journal,twoside]{IEEEtran}
\usepackage{amsmath,amsfonts}
\usepackage{algorithmic}
\usepackage{algorithm}
\usepackage{array}
\usepackage{textcomp}
\usepackage{stfloats}
\usepackage{url}
\usepackage{verbatim}
\usepackage{graphicx}
\usepackage{cite}
\usepackage{import}
\usepackage{tikz}
\usetikzlibrary{patterns,positioning,through,intersections,arrows,arrows.meta,calc,shapes,pgfplots.groupplots,matrix,patterns.meta}
\usepackage{pgfplots}
\pgfplotsset{compat=1.9, table/search path={data}}
\usepgfplotslibrary{fillbetween}
\usepackage{tabularray}
\usepackage{tabularx}
\usepackage{multirow}
\usepackage{colortbl}
\usepackage{booktabs}
\usepackage{siunitx}
\usepackage{amssymb}
\usepackage{subcaption}
\usepackage{textcomp,gensymb}
\usepackage[hidelinks]{hyperref}
\usepackage[nameinlink,capitalise]{cleveref}
\usepackage{url}
\usepackage{orcidlink}

\begin{document}
\title{DUFOMap: Efficient Dynamic Awareness Mapping}

\author{
    Daniel~Duberg\orcidlink{0000-0003-4815-9689},
    Qingwen~Zhang\orcidlink{0000-0002-7882-948X},~\IEEEmembership{Graduate Student Member, IEEE}, \\
    MingKai~Jia\orcidlink{0000-0003-2100-5305},~\IEEEmembership{Graduate Student Member, IEEE}, 
    and~Patric~Jensfelt\orcidlink{0000-0002-1170-7162},~\IEEEmembership{Member,~IEEE}
\thanks{
Manuscript received October 27, 2023; revised February 16, 2024; accepted March 23, 2024. 
This paper was recommended for publication by Editor Moon~Hyungpil upon evaluation of the Associate Editor and Reviewers’ comments. 
This work was supported by Wallenberg AI, Autonomous Systems and Software Program (WASP) including the WASP NEST PerCorSo. 
\textit{(Daniel~Duberg and Qingwen~Zhang are co-first authors.) (Corresponding author: Qingwen~Zhang.)}
}
\thanks{Daniel~Duberg, Qingwen~Zhang, and Patric~Jensfelt are with the Division of Robotics, Perception, and Learning (RPL), KTH Royal Institute of Technology, Stockholm 114 28, Sweden. (email: dduberg@kth.se; qingwen@kth.se; patric@kth.se)}
\thanks{MingKai~Jia is with Robotics Institute, The Hong Kong University of Science and Technology, Hong Kong SAR, China. (email: mjiaab@connect.ust.hk)}
\thanks{Digital Object Identifier (DOI): see top of this page.}
}

\markboth{IEEE ROBOTICS AND AUTOMATION LETTERS. PREPRINT VERSION. ACCEPTED March~2024}%
{D. Duberg, Q. Zhang, M. Jia, P. Jensfelt: DUFOMap Efficient Dynamic Awareness Mapping}


\maketitle

\begin{abstract}
The dynamic nature of the real world is one of the main challenges in robotics. The first step in dealing with it is to detect which parts of the world are dynamic. A typical benchmark task is to create a map that contains only the static part of the world to support, for example, localization and planning. Current solutions are often applied in post-processing, where parameter tuning allows the user to adjust the setting for a specific dataset. In this paper, we propose DUFOMap, a novel dynamic awareness mapping framework designed for efficient online processing. Despite having the same parameter settings for all scenarios, it performs better or is on par with state-of-the-art methods. Ray casting is utilized to identify and classify fully observed empty regions. Since these regions have been observed empty, it follows that anything inside them at another time must be dynamic. Evaluation is carried out in various scenarios, including outdoor environments in KITTI and Argoverse 2, open areas on the KTH campus, and with different sensor types. DUFOMap outperforms the state of the art in terms of accuracy and computational efficiency. (See \href{https://kth-rpl.github.io/dufomap}{https://kth-rpl.github.io/dufomap} for more details.)
\end{abstract}

\begin{IEEEkeywords}
Mapping; Object Detection, Segmentation and Categorization; Robotics and Automation in Construction
\end{IEEEkeywords}

\section{Introduction}
\IEEEPARstart{P}{oint} clouds are a widely used representation in robotics, acquired by sensors such as LiDARs and depth cameras. Point cloud representations also find applications in other domains, e.g., surveying, architecture, and the construction industry. 

Many core components in robotics assume that the environment is static. When this assumption is broken, the robot is often unable to complete its task or, at least, the efficiency is decreased. 
In path planning, dynamics might be misinterpreted as being part of the environment's structure, leading to unnecessarily long or convoluted paths or even failure. 
Dynamic objects incorrectly added to a static map or parts incorrectly removed may also reduce the robustness of localization by introducing ambiguous features or misleading the matching process. To achieve robust operation, the system needs dynamic awareness. 
Today, industrial mapping for global planning and localization is, therefore, typically done offline and under human supervision.

\begin{figure}[t]
    \centering
    \includegraphics[trim=200 40 80 5, clip, width=\linewidth]{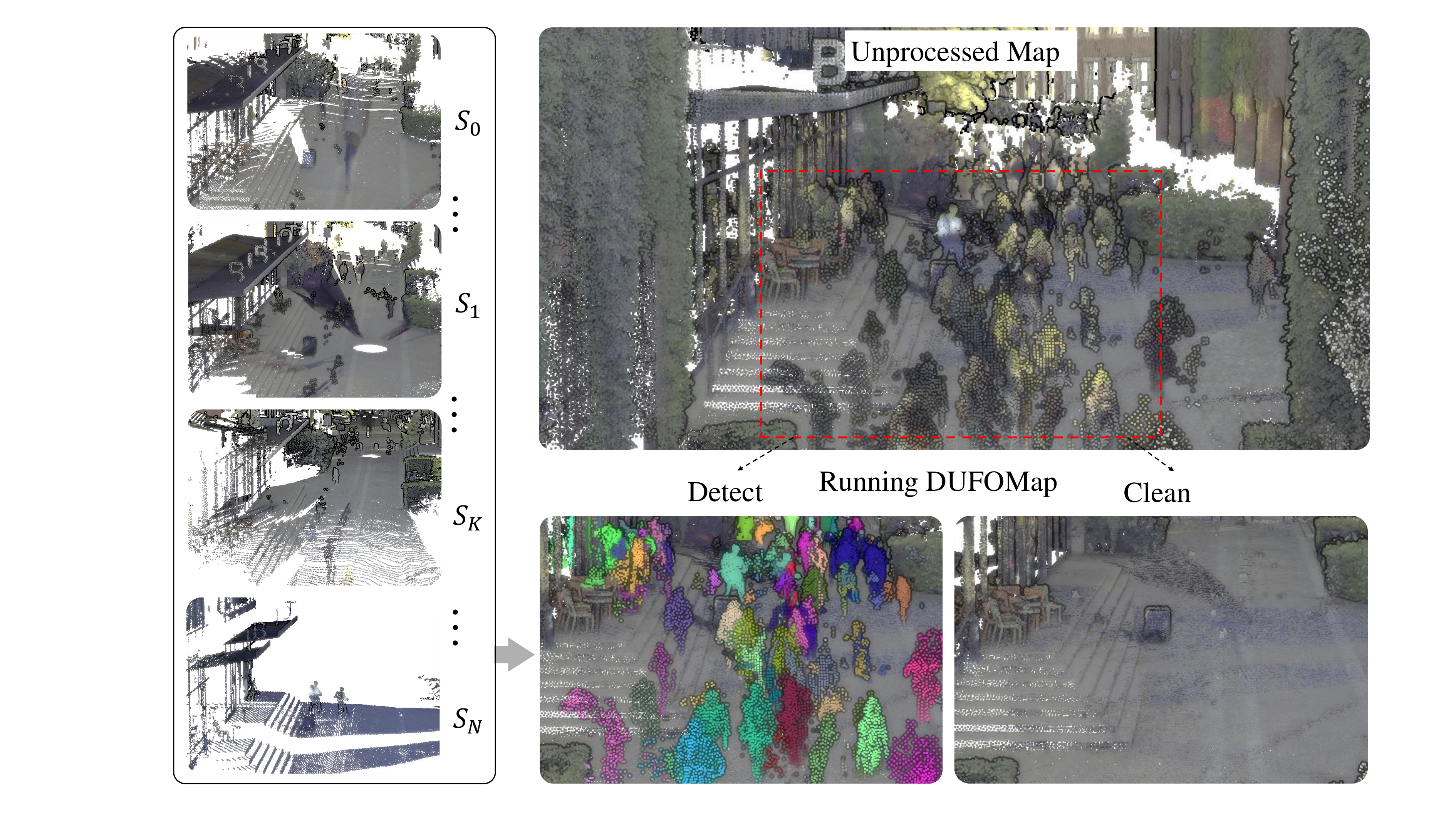}
    \caption{The mapping pipeline integrates all point clouds into a global map, which initially contains numerous dynamic points. The unprocessed map is shown in the upper right. After processing with DUFOMap, the algorithm effectively detects and removes dynamic points, resulting in a clean and refined map suitable for downstream tasks.}
    \label{fig:background}
\end{figure}

An example from surveying where dynamic objects cause problems is shown in \cref{fig:background}. A point cloud model of the built environment is acquired using a 3D laser scanner (Leica-RTC360). By carefully aligning the individual point clouds using artificial reference points, an accurate model can be created. Such a model is often used as a ground truth model for SLAM~\cite{hilti2023challenge, jiao2022fusionportable}. However, as seen in the top right, the quality of the map is severely compromised by the presence of people moving around. 

In this work, we look at classifying points as either static or dynamic. The main evaluation task is map cleaning, where we want to remove the points originating from moving objects and keep the rest in a point cloud map. An example of this is shown at the bottom of \cref{fig:background} where the dynamic points from above have been detected (left) and removed (right).

Several methods are proposed for this. Learning-based methods~\cite{mersch2022ral, 9981210, 9981323,mersch2023ral} require training data and often lack explainability. 
In contrast, methods based on geometric analysis, such as ray casting and visibility~\cite{peopleremover, gskim-2020-iros, lim2021erasor, zhang2023dynamic}, often do not support online execution, as they rely on prior maps for difference calculations, and are computationally and memory expensive. Furthermore, they often require parameter tuning for each new setting.

In some cases, it is important to remove dynamic objects in real-time. For example, local planning cannot only rely on pre-defined maps, as the environment might change during a mission. This rules out methods that rely on first acquiring all sensor data to build a global map before any cleaning happens. 

In this work, we propose DUFOMap, a dynamic awareness method based on UFOMap~\cite{duberg2020ufomap}. The core of the method operates on point clouds which are processed in the voxel structure of UFOMap. Ray casting is used to identify the so-called \emph{void} regions that at some time were empty. The classification of dynamic points can then be done by looking for points that fall into these void regions. Special care is required to account for localization errors and sensor noise. DUFOMap can be used for both offline map cleaning and online detection of dynamic points. In the offline mode, the classification of dynamic points is performed at the end based on all void regions. 

We present extensive experimental validation across multiple datasets, sensors, and scenarios, showing the generality, computational efficiency, and broad usability of DUFOMap. Our approach is open-source at \color{blue}\href{https://github.com/KTH-RPL/dufomap}{https://github.com/KTH-RPL/dufomap}\color{black}.
The main contributions of our work:
\begin{itemize}
    \item We propose a method for detecting dynamics by finding parts of space that has been observed as free taking into account sensor noise and localization errors.
    \item Our method achieves state-of-the-art performance in both offline and online scenarios across different scenarios and sensors.
    \item We demonstrate that our method generalizes in experiments on datasets with five different sensors using the same setting for the method's three parameters.
\end{itemize}

\section{Related Work}
Dynamic awareness algorithms can be broadly categorized into learning-based and geometric analysis methods. 

\subsection{Learning-Based Methods}
Learning-based methods, such as detection and segmentation in point clouds, typically involve deep neural networks and supervised training with labeled datasets. Once trained, these models are capable of inference in real-world scenarios given similar sensor settings.

Mersch~\emph{et al.}~\cite{mersch2022ral}, Sun~\emph{et al.}~\cite{9981210}, and Toyungyernsub~\emph{et al.}~\cite{9981323} develop novel frameworks to extract features and detect dynamic points utilizing spatial and temporal information. Some of these methods use the point cloud format, while others choose to translate point clouds into different representations, such as residual images, to facilitate processing. 
Huang~\emph{et al.}~\cite{huang2022dynamic} propose a novel method for unsupervised point cloud segmentation by jointly learning the latent space representation and a clustering algorithm using a variational autoencoder. Lastly, Khurana~\emph{et al.}~\cite{khurana2022differentiable} use differentiable ray casting to render future occupancy predictions into future LiDAR sweep predictions for learning, allowing geometric occupancy maps to decouple the motion of the environment from the motion of the ego vehicle.

Despite the popularity of learning-based methods, they face several challenges, 
including the need for extensive labeled datasets, handling unbalanced data during training~\cite{zhang2022not} and hard to adapt them to different operation conditions (such as sensors and environments). Additionally, these methods often lack explainability, making it difficult to specify the precise reasons behind poor performance in specific cases. As a result, robustness and generalization continue to be common concerns for learning-based approaches.

\subsection{Geometric Analysis Methods}
Geometric analysis methods do not require labeled data. 
One way to divide these methods (as in~\cite{lim2021erasor}) is into ray-casting and visibility-based methods. Another distinction can be made between methods that operate after all data has been acquired and, therefore, are limited to offline use and methods that can detect dynamic points (and remove them if cleaning is the task) online.

Two of the most popular post-process methods are Removert~\cite{gskim-2020-iros} and ERASOR~\cite{lim2021erasor} with its follow-up ERASOR2~\cite{lim2023erasor2}. They first build a map from all the point clouds and are thus confined to offline operation. Removert~\cite{gskim-2020-iros} projects the map into range images at the location of each query scan. Dynamic points are found based on visibility constraints by comparing query and map range images and using voting. To reduce the number of false positives, range images at decreasing resolution are used to revert dynamic points back to being static.

\begin{figure}[t]
\centering
    \begin{subfigure}[t]{\linewidth}
        \centering
        \includegraphics[trim=340 590 600 110, clip, width=3.0in]{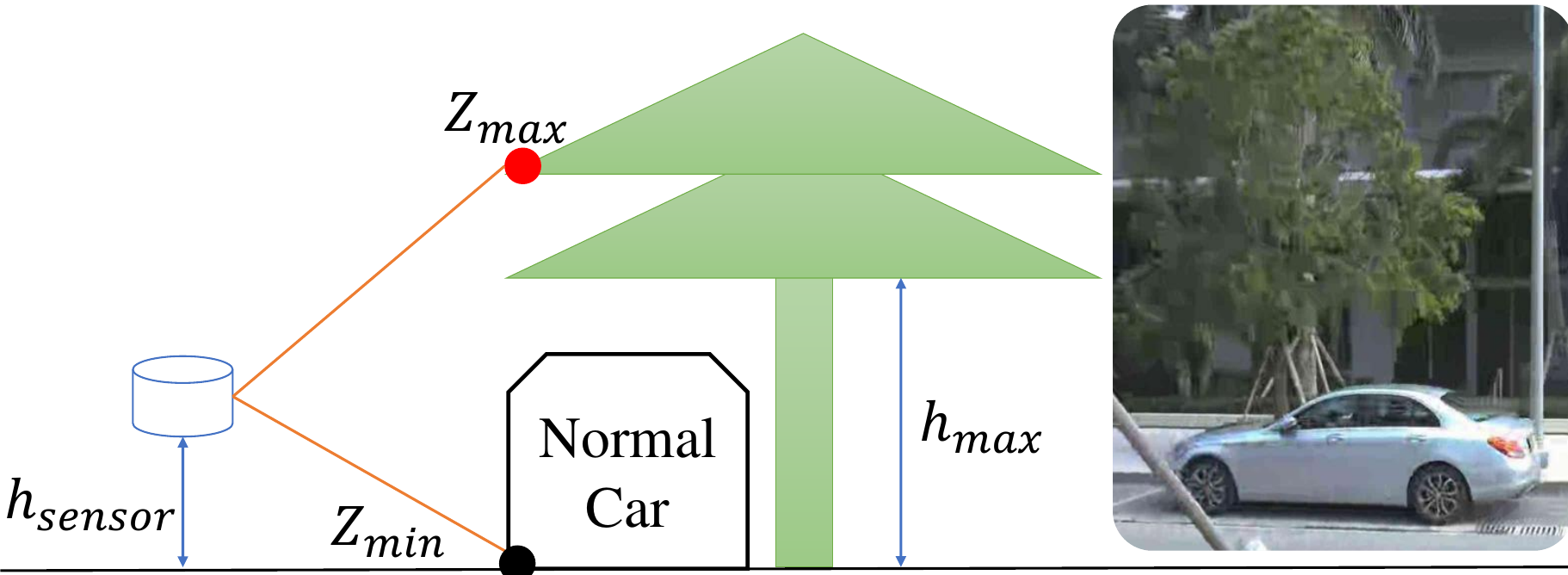}
        \caption{}
    \end{subfigure}
    \hfill
    \begin{subfigure}[t]{\linewidth}
        \centering
        \includegraphics[trim=340 190 600 510, clip, width=3.0in]{figures/height_limitation.pdf}
        \caption{}
    \end{subfigure}
\caption{
Limitations of height threshold. (a) When an object is under, for example, a tree, using a height threshold \( h_{\text{max}} \) typically ignores the tree's highest points (the red point).
(b) However, when choosing a threshold \( h_{\text{max}} \), larger objects, such as a truck, may still have remaining points (purple points).
Two real-world images captured from Argoverse 2 datasets demonstrate these limitations in practice.
}
\label{fig:issue_observe_height}
\end{figure}

To address challenges when the angle between the ray and the ground is small, leading to mislabeling of ground points, Lim~\emph{et al.}~\cite{lim2021erasor} detect dynamic points by assuming that dynamic objects are on the ground. They compare the ratio between the min and max z values in the regions between a query scan and the map. If the ratio is larger than a threshold, then this region contains dynamics, and they remove the full bin. 
This design makes the method sensitive to the sensor height and the parameters that define minimum and maximum height ranges. As a result, each new scenario typically requires a different set of parameters. Furthermore, the method struggles with overhanging objects such as trees, as shown in~\cref{fig:issue_observe_height}.

Occupancy grids, such as OctoMap~\cite{hornung13auro} and UFOMap~\cite{duberg2020ufomap}, use ray casting to update the occupancy values of the voxels in a 3D grid. This results in an estimate of the probability of each voxel being occupied over time. Points that fall into cells with an occupancy probability above some threshold are considered static. Both offline and online operations are possible. For online execution, the classification of a point is performed when the point is acquired. 

The truncated signed distance field (TSDF) is an alternative to occupancy. 
The work closest to ours is Dynablox~\cite{schmid2023dynablox}, which uses Voxblox~\cite{oleynikova2017voxblox} to build a TSDF. So called `ever-free' regions are identified by setting a threshold on the TSDF values of voxels during sequential data updates. It detects dynamics as points falling into these `ever-free' regions.
Dynablox~\cite{schmid2023dynablox} runs online in a sliding window.

\subsection{Summary}
Both geometric analysis and learning-based approaches have their advantages and limitations. Robustness and generalization are common issues in learning-based methods; most importantly, they require large pre-labeled training datasets, which can be labor-intensive and time-consuming to create. State-of-the-art geometric analysis methods are typically operated offline and can therefore afford to have parameters that might need changes per scenario. 
Our proposed method, DUFOMap, is designed for online dynamic awareness mapping where tuning parameters for different scenarios is not possible, but, as will be demonstrated, outperforms offline methods.

\section{Method}
Following earlier work~\cite{lindstrom2001detecting}, the underlying idea behind our proposed approach is to identify empty regions of space, rather than directly identify dynamic regions. The key insight is that if a region has been observed as empty at one time, points observed inside this region at another time have to be dynamic. We call a region that has been observed empty at least once a \emph{void} region. 

DUFOMap, being an extension of UFOMap~\cite{duberg2020ufomap}, discretizes the world into voxels. Each voxel contains a flag \( i_{\text{void}} \) indicating whether the voxel has been observed empty at some time. Initially, \( i_{\text{void}} \) is \emph{false} for each voxel; the \( i_{\text{void}} \) is changed to \emph{true} when the voxel has been observed empty. 
A point in space is classified as static or dynamic by looking at the \( i_{\text{void}} \) flag for the corresponding voxel.

In this work, we assume that the input is pairs of sensor poses and point clouds (see~\cref{fig:step_1}). The sensor pose is the position and orientation of the sensor in the world frame. 
There are no assumptions about the structure of the point clouds allowing for, for example, non-repetitive scanning patterns.

\subsection{Classifying Void Regions}
The classification of the void regions differs from the \emph{free space} used in, for example, occupancy grids. Occupancy grids often employ a probabilistic model that updates a region based on all observations of that region. In an occupancy grid, the same part of the space can switch between being identified as free and occupied. In our work, a void region is classified as such from a single point cloud observation. 

Classifying void regions from single observations, rather than accumulating evidence over time, means that a region can be classified quickly; but also that great care must be taken to prevent misclassifications. 
Each region is represented by a voxel, which is classified as void if it has been observed to be completely empty at least once.

\begin{figure}[t]
    \centering
    \begin{subfigure}[t]{0.49\linewidth}
        \centering
        \resizebox{1.0\linewidth}{!}{
            \begin{tikzpicture}
    \definecolor{freecolor}{RGB}{178,171,210}
    \definecolor{occupiedcolor}{RGB}{230,97,1}
    \definecolor{raycolor}{RGB}{253,184,99}
    \definecolor{sensorcolor}{RGB}{94,60,153}
    \definecolor{hitcolor}{RGB}{110,110,110}
    \definecolor{compensatecolor}{RGB}{0,255,0}

    \coordinate (sensor) at (1.4,5.7);
    
    \fill[freecolor] (1,5) rectangle +(1,1);
    \fill[freecolor] (2,5) rectangle +(1,1);
    \fill[freecolor] (3,5) rectangle +(1,1);
    \fill[freecolor] (4,5) rectangle +(1,1);
    \fill[freecolor] (5,5) rectangle +(1,1);
    \fill[freecolor] (6,5) rectangle +(1,1);
    \fill[freecolor] (7,5) rectangle +(1,1);
    \fill[freecolor] (8,5) rectangle +(1,1);
    \fill[freecolor] (9,5) rectangle +(1,1);
    \fill[freecolor] (10,5) rectangle +(1,1);
    \fill[freecolor] (11,5) rectangle +(1,1);
    
    \fill[freecolor] (3,4) rectangle +(1,1);
    \fill[freecolor] (4,4) rectangle +(1,1);
    \fill[freecolor] (5,4) rectangle +(1,1);
    \fill[freecolor] (6,4) rectangle +(1,1);
    \fill[freecolor] (7,4) rectangle +(1,1);
    \fill[freecolor] (8,4) rectangle +(1,1);
    \fill[freecolor] (9,4) rectangle +(1,1);
    \fill[freecolor] (10,4) rectangle +(1,1);
    \fill[freecolor] (11,4) rectangle +(1,1);
    \fill[freecolor] (13,4) rectangle +(1,1);
    \fill[freecolor] (14,4) rectangle +(1,1);

    \fill[freecolor] (6,3) rectangle +(1,1);
    \fill[freecolor] (7,3) rectangle +(1,1);
    \fill[freecolor] (8,3) rectangle +(1,1);
    \fill[freecolor] (9,3) rectangle +(1,1);
    \fill[freecolor] (11,3) rectangle +(1,1);
    
    \fill[freecolor] (8,2) rectangle +(1,1);
    \fill[freecolor] (9,2) rectangle +(1,1);
    
    \fill[freecolor] (2,6) rectangle +(1,1);
    \fill[freecolor] (3,6) rectangle +(1,1);
    \fill[freecolor] (4,6) rectangle +(1,1);
    \fill[freecolor] (5,6) rectangle +(1,1);
    \fill[freecolor] (6,6) rectangle +(1,1);
    \fill[freecolor] (7,6) rectangle +(1,1);
    \fill[freecolor] (8,6) rectangle +(1,1);
    \fill[freecolor] (9,6) rectangle +(1,1);
    \fill[freecolor] (10,6) rectangle +(1,1);
    \fill[freecolor] (11,6) rectangle +(1,1);
    
    \fill[freecolor] (5,7) rectangle +(1,1);
    \fill[freecolor] (6,7) rectangle +(1,1);
    \fill[freecolor] (7,7) rectangle +(1,1);
    \fill[freecolor] (8,7) rectangle +(1,1);
    \fill[freecolor] (9,7) rectangle +(1,1);
    \fill[freecolor] (11,7) rectangle +(1,1);
    
    \fill[freecolor] (7,8) rectangle +(1,1);
    \fill[freecolor] (8,8) rectangle +(1,1);
    \fill[freecolor] (9,8) rectangle +(1,1);
    
    \fill[freecolor] (10,9) rectangle +(1,1);
    
    \fill[hitcolor] (11,9) rectangle +(1,1);
    \fill[hitcolor] (10,8) rectangle +(1,1);
    \fill[hitcolor] (10,7) rectangle +(1,1);
    \fill[hitcolor] (12,7) rectangle +(1,1);
    \fill[hitcolor] (12,6) rectangle +(1,1);
    \fill[hitcolor] (12,5) rectangle +(1,1);
    \fill[hitcolor] (12,4) rectangle +(1,1);
    \fill[hitcolor] (15,4) rectangle +(1,1);
    \fill[hitcolor] (12,3) rectangle +(1,1);
    \fill[hitcolor] (10,3) rectangle +(1,1);
    \fill[hitcolor] (10,2) rectangle +(1,1);
    
    \draw[step=4cm,gray,ultra thick] (0,2) grid (16,10);
    \draw[step=2cm,gray,thick] (0,2) grid (16,10);
    \draw[step=1cm,gray,thin] (0,2) grid (16,10);
    
    \draw[raycolor,ultra thick] (1.4,5.7) -- ++(-20:10cm) coordinate (hit1);
    \draw[raycolor,ultra thick] (1.4,5.7) -- ++(-18:9.9cm) coordinate (hit3);
    \draw[raycolor,ultra thick] (1.4,5.7) -- ++(-16:9.9cm) coordinate (hit5);
    \draw[raycolor,ultra thick] (1.4,5.7) -- ++(-14:9.7cm) coordinate (hit7);
    \draw[raycolor,ultra thick] (1.4,5.7) -- ++(-12:9.5cm) coordinate (hit9);
    
    \draw[raycolor,ultra thick] (1.4,5.7) -- ++(-10:10.9cm) coordinate (hit11);
    \draw[raycolor,ultra thick] (1.4,5.7) -- ++(-8:10.95cm) coordinate (hit13);
    \draw[raycolor,ultra thick] (1.4,5.7) -- ++(-6:10.9cm) coordinate (hit15);
    \draw[raycolor,ultra thick] (1.4,5.7) -- ++(-4:14cm) coordinate (hit17);
    \draw[raycolor,ultra thick] (1.4,5.7) -- ++(-2:11cm) coordinate (hit19);
    \draw[raycolor,ultra thick] (1.4,5.7) -- ++(0:11cm) coordinate (hit21);
    \draw[raycolor,ultra thick] (1.4,5.7) -- ++(2:11cm) coordinate (hit23);
    \draw[raycolor,ultra thick] (1.4,5.7) -- ++(4:11cm) coordinate (hit25);
    \draw[raycolor,ultra thick] (1.4,5.7) -- ++(6:11cm) coordinate (hit27);
    \draw[raycolor,ultra thick] (1.4,5.7) -- ++(8:11.03cm) coordinate (hit29);
    
    \draw[raycolor,ultra thick] (1.4,5.7) -- ++(10:9.0cm) coordinate (hit31);
    \draw[raycolor,ultra thick] (1.4,5.7) -- ++(12:9.0cm) coordinate (hit33);
    \draw[raycolor,ultra thick] (1.4,5.7) -- ++(14:9.08cm) coordinate (hit35);
    \draw[raycolor,ultra thick] (1.4,5.7) -- ++(16:9.02cm) coordinate (hit37);
    \draw[raycolor,ultra thick] (1.4,5.7) -- ++(18:9.1cm) coordinate (hit39);
    \draw[raycolor,ultra thick] (1.4,5.7) -- ++(20:11cm) coordinate (hit41);
    
    \draw[fill,occupiedcolor] (hit1) circle (0.1);
    \draw[fill,occupiedcolor] (hit3) circle (0.1);
    \draw[fill,occupiedcolor] (hit5) circle (0.1);
    \draw[fill,occupiedcolor] (hit7) circle (0.1);
    \draw[fill,occupiedcolor] (hit9) circle (0.1);
    \draw[fill,occupiedcolor] (hit11) circle (0.1);
    \draw[fill,occupiedcolor] (hit13) circle (0.1);
    \draw[fill,occupiedcolor] (hit15) circle (0.1);
    \draw[fill,occupiedcolor] (hit17) circle (0.1);
    \draw[fill,occupiedcolor] (hit19) circle (0.1);
    \draw[fill,occupiedcolor] (hit21) circle (0.1);
    \draw[fill,occupiedcolor] (hit23) circle (0.1);
    \draw[fill,occupiedcolor] (hit25) circle (0.1);
    \draw[fill,occupiedcolor] (hit27) circle (0.1);
    \draw[fill,occupiedcolor] (hit29) circle (0.1);
    \draw[fill,occupiedcolor] (hit31) circle (0.1);
    \draw[fill,occupiedcolor] (hit33) circle (0.1);
    \draw[fill,occupiedcolor] (hit35) circle (0.1);
    \draw[fill,occupiedcolor] (hit37) circle (0.1);
    \draw[fill,occupiedcolor] (hit39) circle (0.1);
    \draw[fill,occupiedcolor] (hit41) circle (0.1);
    
    \path[draw, thick] (sensor) --++ (0.45,0.3) --++ (0, -0.6) -- cycle;
\end{tikzpicture}
        }
        \caption{}
        \label{fig:step_1}
    \end{subfigure}
    \hfill
    \begin{subfigure}[t]{0.49\linewidth}
        \centering
        \resizebox{1.0\linewidth}{!}{
            \begin{tikzpicture}
    \definecolor{freecolor}{RGB}{178,171,210}
    \definecolor{occupiedcolor}{RGB}{230,97,1}
    \definecolor{raycolor}{RGB}{253,184,99}
    \definecolor{sensorcolor}{RGB}{94,60,153}
    \definecolor{hitcolor}{RGB}{110,110,110}
    \definecolor{compensatecolor}{RGB}{0,255,0}
    \definecolor{somethingcolor}{RGB}{0,255,255}
    \definecolor{deffreecolor}{RGB}{255,0,0}

    \coordinate (sensor) at (1.4,5.7);
    
    \fill[freecolor] (1,5) rectangle +(1,1);
    \fill[freecolor] (2,5) rectangle +(1,1);
    \fill[freecolor] (3,5) rectangle +(1,1);
    \fill[freecolor] (4,5) rectangle +(1,1);
    \fill[freecolor] (5,5) rectangle +(1,1);
    \fill[freecolor] (6,5) rectangle +(1,1);
    \fill[freecolor] (7,5) rectangle +(1,1);
    \fill[freecolor] (8,5) rectangle +(1,1);
    \fill[freecolor] (9,5) rectangle +(1,1);
    \fill[freecolor] (10,5) rectangle +(1,1);
    \fill[freecolor] (11,5) rectangle +(1,1);
    
    \fill[freecolor] (3,4) rectangle +(1,1);
    \fill[freecolor] (4,4) rectangle +(1,1);
    \fill[freecolor] (5,4) rectangle +(1,1);
    \fill[freecolor] (6,4) rectangle +(1,1);
    \fill[freecolor] (7,4) rectangle +(1,1);
    \fill[freecolor] (8,4) rectangle +(1,1);
    \fill[freecolor] (9,4) rectangle +(1,1);
    \fill[freecolor] (10,4) rectangle +(1,1);
    \fill[freecolor] (11,4) rectangle +(1,1);
    \fill[freecolor] (13,4) rectangle +(1,1);
    \fill[freecolor] (14,4) rectangle +(1,1);

    \fill[freecolor] (6,3) rectangle +(1,1);
    \fill[freecolor] (7,3) rectangle +(1,1);
    \fill[freecolor] (8,3) rectangle +(1,1);
    \fill[freecolor] (9,3) rectangle +(1,1);
    \fill[freecolor] (11,3) rectangle +(1,1);
    
    \fill[freecolor] (8,2) rectangle +(1,1);
    \fill[freecolor] (9,2) rectangle +(1,1);
    
    \fill[freecolor] (2,6) rectangle +(1,1);
    \fill[freecolor] (3,6) rectangle +(1,1);
    \fill[freecolor] (4,6) rectangle +(1,1);
    \fill[freecolor] (5,6) rectangle +(1,1);
    \fill[freecolor] (6,6) rectangle +(1,1);
    \fill[freecolor] (7,6) rectangle +(1,1);
    \fill[freecolor] (8,6) rectangle +(1,1);
    \fill[freecolor] (9,6) rectangle +(1,1);
    \fill[freecolor] (10,6) rectangle +(1,1);
    \fill[freecolor] (11,6) rectangle +(1,1);
    
    \fill[freecolor] (5,7) rectangle +(1,1);
    \fill[freecolor] (6,7) rectangle +(1,1);
    \fill[freecolor] (7,7) rectangle +(1,1);
    \fill[freecolor] (8,7) rectangle +(1,1);
    \fill[freecolor] (9,7) rectangle +(1,1);
    \fill[freecolor] (11,7) rectangle +(1,1);
    
    \fill[freecolor] (7,8) rectangle +(1,1);
    \fill[freecolor] (8,8) rectangle +(1,1);
    \fill[freecolor] (9,8) rectangle +(1,1);
    
    \fill[freecolor] (10,9) rectangle +(1,1);
    
    \fill[hitcolor] (11,9) rectangle +(1,1);
    \fill[hitcolor] (10,8) rectangle +(1,1);
    \fill[hitcolor] (10,7) rectangle +(1,1);
    \fill[hitcolor] (12,7) rectangle +(1,1);
    \fill[hitcolor] (12,6) rectangle +(1,1);
    \fill[hitcolor] (12,5) rectangle +(1,1);
    \fill[hitcolor] (12,4) rectangle +(1,1);
    \fill[hitcolor] (15,4) rectangle +(1,1);
    \fill[hitcolor] (12,3) rectangle +(1,1);
    \fill[hitcolor] (10,3) rectangle +(1,1);
    \fill[hitcolor] (10,2) rectangle +(1,1);
    
    \draw[step=4cm,gray,ultra thick] (0,2) grid (16,10);
    \draw[step=2cm,gray,thick] (0,2) grid (16,10);
    \draw[step=1cm,gray,thin] (0,2) grid (16,10);
    
    \fill[deffreecolor] (4,5) rectangle +(1,1);
    \fill[deffreecolor] (5,5) rectangle +(1,1);
    \fill[deffreecolor] (6,5) rectangle +(1,1);
    \fill[deffreecolor] (7,5) rectangle +(1,1);
    \fill[deffreecolor] (8,5) rectangle +(1,1);
    \fill[deffreecolor] (9,5) rectangle +(1,1);
    \fill[deffreecolor] (10,5) rectangle +(1,1);
    \fill[deffreecolor] (11,5) rectangle +(1,1);

    \fill[deffreecolor] (6,6) rectangle +(1,1);
    \fill[deffreecolor] (7,6) rectangle +(1,1);
    \fill[deffreecolor] (8,6) rectangle +(1,1);
    \fill[deffreecolor] (9,6) rectangle +(1,1);
    \fill[deffreecolor] (10,6) rectangle +(1,1);
    \fill[deffreecolor] (11,6) rectangle +(1,1);
    
    \fill[deffreecolor] (8,7) rectangle +(1,1);
    \fill[deffreecolor] (9,7) rectangle +(1,1);

    \fill[deffreecolor] (7,4) rectangle +(1,1);
    \fill[deffreecolor] (8,4) rectangle +(1,1);
    \fill[deffreecolor] (9,4) rectangle +(1,1);
    \fill[deffreecolor] (10,4) rectangle +(1,1);
    \fill[deffreecolor] (11,4) rectangle +(1,1);
    
    \fill[deffreecolor] (9,3) rectangle +(1,1);
    
\end{tikzpicture}
        }
        \caption{}
        \label{fig:step_2}
    \end{subfigure}
    \caption{Example of point cloud integration in DUFOMap (shown as a single slice of the 3D grid). (a) From the sensor position, the triangle to the left, ray casting (orange lines) is performed for each point (orange dot) in the point cloud. All cells intersecting a ray are marked as intersected (purple), and the cells where a point falls within are marked as hit (gray). Unknown cells are white. (b) Cells that are intersected and surrounded exclusively by other intersected or hit cells are classified as void regions (red).}
    \label{fig:method}
\end{figure}
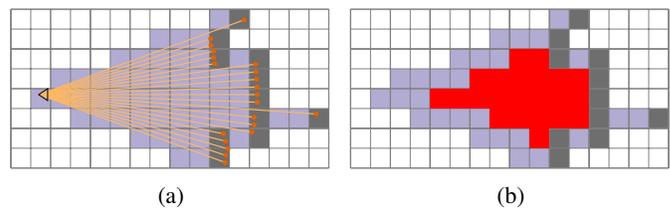

To identify candidate void voxels, ray casting is performed from the sensor position to each point in the point cloud. After casting the rays, a voxel can be in one of three states (see~\cref{fig:step_2}): \emph{hit} (gray) if a point fell inside the voxel, \emph{intersected} (purple) if the voxel was intersected by a ray and no point fell into it, and \emph{unknown} (white) otherwise. The candidate void voxels are those in the intersected state.

However, a single ray intersecting a voxel does not mean that the whole voxel has been observed; therefore, it is not guaranteed to be a void voxel based exclusively on this. To this end, we propose looking at the neighboring voxels. We define the voxel in question to be fully observed given the capabilities of the sensor in use if all 26 surrounding voxels in 3D (or 8 in 2D as in \cref{fig:method,fig:integration_comparision}) are also in the intersected or hit state (illustrated in red in~\cref{fig:step_2}). Given this requirement, the voxels at the borders of the volume spanned by the point cloud cannot be confirmed to be void, as they neighbor unknown voxels.

\subsection{Dealing With the Real World}
\begin{figure}[t]
    \centering
    \begin{subfigure}[t]{0.49\linewidth}
        \centering
        \resizebox{1.0\linewidth}{!}{%
            \import{figures/tikz/}{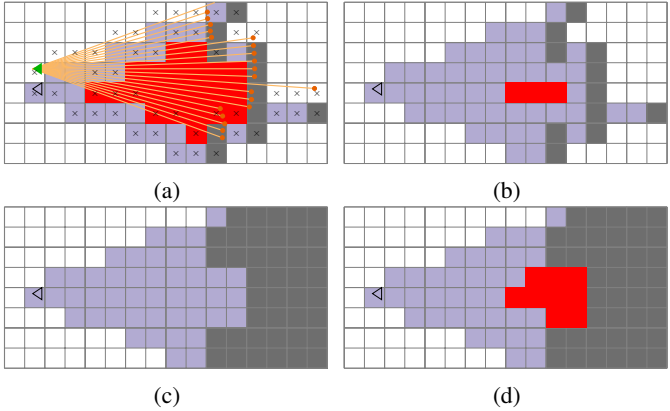}%
        }%
        \caption{}
        \label{fig:method_4}
    \end{subfigure}%
    \hfill
    \begin{subfigure}[t]{0.49\linewidth}
        \centering
        \resizebox{1.0\linewidth}{!}{%
            \begin{tikzpicture}
    \definecolor{freecolor}{RGB}{178,171,210}
    \definecolor{occupiedcolor}{RGB}{230,97,1}
    \definecolor{raycolor}{RGB}{253,184,99}
    \definecolor{sensorcolor}{RGB}{94,60,153}
    \definecolor{hitcolor}{RGB}{110,110,110}
    \definecolor{compensatecolor}{RGB}{0,255,0}
    \definecolor{somethingcolor}{RGB}{0,255,255}
    \definecolor{deffreecolor}{RGB}{255,0,0}

    \coordinate (sensor) at (1.4,5.7);
    
    \fill[freecolor] (1,5) rectangle +(1,1);
    \fill[freecolor] (2,5) rectangle +(1,1);
    \fill[freecolor] (3,5) rectangle +(1,1);
    \fill[freecolor] (4,5) rectangle +(1,1);
    \fill[freecolor] (5,5) rectangle +(1,1);
    \fill[freecolor] (6,5) rectangle +(1,1);
    \fill[freecolor] (7,5) rectangle +(1,1);
    \fill[freecolor] (8,5) rectangle +(1,1);
    \fill[freecolor] (9,5) rectangle +(1,1);
    \fill[freecolor] (10,5) rectangle +(1,1);
    \fill[freecolor] (11,5) rectangle +(1,1);
    
    \fill[freecolor] (3,4) rectangle +(1,1);
    \fill[freecolor] (4,4) rectangle +(1,1);
    \fill[freecolor] (5,4) rectangle +(1,1);
    \fill[freecolor] (6,4) rectangle +(1,1);
    \fill[freecolor] (7,4) rectangle +(1,1);
    \fill[freecolor] (8,4) rectangle +(1,1);
    \fill[freecolor] (9,4) rectangle +(1,1);
    \fill[freecolor] (10,4) rectangle +(1,1);
    \fill[freecolor] (11,4) rectangle +(1,1);
    \fill[freecolor] (13,4) rectangle +(1,1);
    \fill[freecolor] (14,4) rectangle +(1,1);

    \fill[freecolor] (6,3) rectangle +(1,1);
    \fill[freecolor] (7,3) rectangle +(1,1);
    \fill[freecolor] (8,3) rectangle +(1,1);
    \fill[freecolor] (9,3) rectangle +(1,1);
    \fill[freecolor] (11,3) rectangle +(1,1);
    
    \fill[freecolor] (8,2) rectangle +(1,1);
    \fill[freecolor] (9,2) rectangle +(1,1);
    
    \fill[freecolor] (2,6) rectangle +(1,1);
    \fill[freecolor] (3,6) rectangle +(1,1);
    \fill[freecolor] (4,6) rectangle +(1,1);
    \fill[freecolor] (5,6) rectangle +(1,1);
    \fill[freecolor] (6,6) rectangle +(1,1);
    \fill[freecolor] (7,6) rectangle +(1,1);
    \fill[freecolor] (8,6) rectangle +(1,1);
    \fill[freecolor] (9,6) rectangle +(1,1);
    \fill[freecolor] (10,6) rectangle +(1,1);
    \fill[freecolor] (11,6) rectangle +(1,1);
    
    \fill[freecolor] (5,7) rectangle +(1,1);
    \fill[freecolor] (6,7) rectangle +(1,1);
    \fill[freecolor] (7,7) rectangle +(1,1);
    \fill[freecolor] (8,7) rectangle +(1,1);
    \fill[freecolor] (9,7) rectangle +(1,1);
    \fill[freecolor] (11,7) rectangle +(1,1);
    
    \fill[freecolor] (7,8) rectangle +(1,1);
    \fill[freecolor] (8,8) rectangle +(1,1);
    \fill[freecolor] (9,8) rectangle +(1,1);
    
    \fill[freecolor] (10,9) rectangle +(1,1);
    
    \fill[hitcolor] (11,9) rectangle +(1,1);
    \fill[hitcolor] (10,8) rectangle +(1,1);
    \fill[hitcolor] (10,7) rectangle +(1,1);
    \fill[hitcolor] (12,7) rectangle +(1,1);
    \fill[hitcolor] (12,6) rectangle +(1,1);
    \fill[hitcolor] (12,5) rectangle +(1,1);
    \fill[hitcolor] (12,4) rectangle +(1,1);
    \fill[hitcolor] (15,4) rectangle +(1,1);
    \fill[hitcolor] (12,3) rectangle +(1,1);
    \fill[hitcolor] (10,3) rectangle +(1,1);
    \fill[hitcolor] (10,2) rectangle +(1,1);
    
    \draw[step=4cm,gray,ultra thick] (0,2) grid (16,10);
    \draw[step=2cm,gray,thick] (0,2) grid (16,10);
    \draw[step=1cm,gray,thin] (0,2) grid (16,10);
    
    \fill[deffreecolor] (8,5) rectangle +(1,1);
    \fill[deffreecolor] (9,5) rectangle +(1,1);
    \fill[deffreecolor] (10,5) rectangle +(1,1);
    
    \path[draw, thick] (sensor) -- ++(0.45,0.3) -- ++(0, -0.6) -- cycle;
\end{tikzpicture}%
        }%
        \caption{}
        \label{fig:method_7}
    \end{subfigure}%
    \vfill
    \begin{subfigure}[t]{0.49\linewidth}
        \centering
        \resizebox{1.0\linewidth}{!}{%
            \begin{tikzpicture}
    \definecolor{freecolor}{RGB}{178,171,210}
    \definecolor{occupiedcolor}{RGB}{230,97,1}
    \definecolor{raycolor}{RGB}{253,184,99}
    \definecolor{sensorcolor}{RGB}{94,60,153}
    \definecolor{hitcolor}{RGB}{110,110,110}
    \definecolor{compensatecolor}{RGB}{0,255,0}
    \definecolor{somethingcolor}{RGB}{0,255,255}
    \definecolor{deffreecolor}{RGB}{255,0,0}

    \coordinate (sensor) at (1.4,5.7);
    
    \fill[freecolor] (1,5) rectangle +(1,1);
    \fill[freecolor] (2,5) rectangle +(1,1);
    \fill[freecolor] (3,5) rectangle +(1,1);
    \fill[freecolor] (4,5) rectangle +(1,1);
    \fill[freecolor] (5,5) rectangle +(1,1);
    \fill[freecolor] (6,5) rectangle +(1,1);
    \fill[freecolor] (7,5) rectangle +(1,1);
    \fill[freecolor] (8,5) rectangle +(1,1);
    \fill[freecolor] (9,5) rectangle +(1,1);
    \fill[freecolor] (10,5) rectangle +(1,1);
    \fill[freecolor] (11,5) rectangle +(1,1);
    
    \fill[freecolor] (3,4) rectangle +(1,1);
    \fill[freecolor] (4,4) rectangle +(1,1);
    \fill[freecolor] (5,4) rectangle +(1,1);
    \fill[freecolor] (6,4) rectangle +(1,1);
    \fill[freecolor] (7,4) rectangle +(1,1);
    \fill[freecolor] (8,4) rectangle +(1,1);
    \fill[freecolor] (9,4) rectangle +(1,1);
    \fill[freecolor] (10,4) rectangle +(1,1);
    \fill[freecolor] (11,4) rectangle +(1,1);
    \fill[freecolor] (13,4) rectangle +(1,1);
    \fill[freecolor] (14,4) rectangle +(1,1);

    \fill[freecolor] (6,3) rectangle +(1,1);
    \fill[freecolor] (7,3) rectangle +(1,1);
    \fill[freecolor] (8,3) rectangle +(1,1);
    \fill[freecolor] (9,3) rectangle +(1,1);
    \fill[freecolor] (11,3) rectangle +(1,1);
    
    \fill[freecolor] (8,2) rectangle +(1,1);
    \fill[freecolor] (9,2) rectangle +(1,1);
    
    \fill[freecolor] (2,6) rectangle +(1,1);
    \fill[freecolor] (3,6) rectangle +(1,1);
    \fill[freecolor] (4,6) rectangle +(1,1);
    \fill[freecolor] (5,6) rectangle +(1,1);
    \fill[freecolor] (6,6) rectangle +(1,1);
    \fill[freecolor] (7,6) rectangle +(1,1);
    \fill[freecolor] (8,6) rectangle +(1,1);
    \fill[freecolor] (9,6) rectangle +(1,1);
    \fill[freecolor] (10,6) rectangle +(1,1);
    \fill[freecolor] (11,6) rectangle +(1,1);
    
    \fill[freecolor] (5,7) rectangle +(1,1);
    \fill[freecolor] (6,7) rectangle +(1,1);
    \fill[freecolor] (7,7) rectangle +(1,1);
    \fill[freecolor] (8,7) rectangle +(1,1);
    \fill[freecolor] (9,7) rectangle +(1,1);
    \fill[freecolor] (11,7) rectangle +(1,1);
    
    \fill[freecolor] (7,8) rectangle +(1,1);
    \fill[freecolor] (8,8) rectangle +(1,1);
    \fill[freecolor] (9,8) rectangle +(1,1);
    
    \fill[freecolor] (10,9) rectangle +(1,1);
    
    \fill[hitcolor] (11,9) rectangle +(1,1);
    \fill[hitcolor] (12,9) rectangle +(1,1);
    \fill[hitcolor] (13,9) rectangle +(1,1);
    \fill[hitcolor] (14,9) rectangle +(1,1);
    \fill[hitcolor] (15,9) rectangle +(1,1);
    \fill[hitcolor] (10,8) rectangle +(1,1);
    \fill[hitcolor] (11,8) rectangle +(1,1);
    \fill[hitcolor] (12,8) rectangle +(1,1);
    \fill[hitcolor] (13,8) rectangle +(1,1);
    \fill[hitcolor] (14,8) rectangle +(1,1);
    \fill[hitcolor] (15,8) rectangle +(1,1);
    \fill[hitcolor] (10,7) rectangle +(1,1);
    \fill[hitcolor] (11,7) rectangle +(1,1);
    \fill[hitcolor] (12,7) rectangle +(1,1);
    \fill[hitcolor] (13,7) rectangle +(1,1);
    \fill[hitcolor] (14,7) rectangle +(1,1);
    \fill[hitcolor] (15,7) rectangle +(1,1);
    \fill[hitcolor] (12,6) rectangle +(1,1);
    \fill[hitcolor] (13,6) rectangle +(1,1);
    \fill[hitcolor] (14,6) rectangle +(1,1);
    \fill[hitcolor] (15,6) rectangle +(1,1);
    \fill[hitcolor] (12,5) rectangle +(1,1);
    \fill[hitcolor] (13,5) rectangle +(1,1);
    \fill[hitcolor] (14,5) rectangle +(1,1);
    \fill[hitcolor] (15,5) rectangle +(1,1);
    \fill[hitcolor] (12,4) rectangle +(1,1);
    \fill[hitcolor] (13,4) rectangle +(1,1);
    \fill[hitcolor] (14,4) rectangle +(1,1);
    \fill[hitcolor] (15,4) rectangle +(1,1);
    \fill[hitcolor] (10,3) rectangle +(1,1);
    \fill[hitcolor] (11,3) rectangle +(1,1);
    \fill[hitcolor] (12,3) rectangle +(1,1);
    \fill[hitcolor] (13,3) rectangle +(1,1);
    \fill[hitcolor] (14,3) rectangle +(1,1);
    \fill[hitcolor] (15,3) rectangle +(1,1);
    \fill[hitcolor] (10,2) rectangle +(1,1);
    \fill[hitcolor] (11,2) rectangle +(1,1);
    \fill[hitcolor] (12,2) rectangle +(1,1);
    \fill[hitcolor] (13,2) rectangle +(1,1);
    \fill[hitcolor] (14,2) rectangle +(1,1);
    \fill[hitcolor] (15,2) rectangle +(1,1);
    
    \draw[step=4cm,gray,ultra thick] (0,2) grid (16,10);
    \draw[step=2cm,gray,thick] (0,2) grid (16,10);
    \draw[step=1cm,gray,thin] (0,2) grid (16,10);

    \path[draw, thick] (sensor) -- ++(0.45,0.3) -- ++(0, -0.6) -- cycle;
\end{tikzpicture}%
        }%
        \caption{}
        \label{fig:method_8}
    \end{subfigure}%
    \hfill
    \begin{subfigure}[t]{0.49\linewidth}
        \centering
        \resizebox{1.0\linewidth}{!}{%
            \begin{tikzpicture}
    \definecolor{freecolor}{RGB}{178,171,210}
    \definecolor{occupiedcolor}{RGB}{230,97,1}
    \definecolor{raycolor}{RGB}{253,184,99}
    \definecolor{sensorcolor}{RGB}{94,60,153}
    \definecolor{hitcolor}{RGB}{110,110,110}
    \definecolor{compensatecolor}{RGB}{0,255,0}
    \definecolor{somethingcolor}{RGB}{0,255,255}
    \definecolor{deffreecolor}{RGB}{255,0,0}

    \coordinate (sensor) at (1.4,5.7);
    
    \fill[freecolor] (1,5) rectangle +(1,1);
    \fill[freecolor] (2,5) rectangle +(1,1);
    \fill[freecolor] (3,5) rectangle +(1,1);
    \fill[freecolor] (4,5) rectangle +(1,1);
    \fill[freecolor] (5,5) rectangle +(1,1);
    \fill[freecolor] (6,5) rectangle +(1,1);
    \fill[freecolor] (7,5) rectangle +(1,1);
    \fill[freecolor] (8,5) rectangle +(1,1);
    \fill[freecolor] (9,5) rectangle +(1,1);
    \fill[freecolor] (10,5) rectangle +(1,1);
    \fill[freecolor] (11,5) rectangle +(1,1);
    
    \fill[freecolor] (3,4) rectangle +(1,1);
    \fill[freecolor] (4,4) rectangle +(1,1);
    \fill[freecolor] (5,4) rectangle +(1,1);
    \fill[freecolor] (6,4) rectangle +(1,1);
    \fill[freecolor] (7,4) rectangle +(1,1);
    \fill[freecolor] (8,4) rectangle +(1,1);
    \fill[freecolor] (9,4) rectangle +(1,1);
    \fill[freecolor] (10,4) rectangle +(1,1);
    \fill[freecolor] (11,4) rectangle +(1,1);
    \fill[freecolor] (13,4) rectangle +(1,1);
    \fill[freecolor] (14,4) rectangle +(1,1);

    \fill[freecolor] (6,3) rectangle +(1,1);
    \fill[freecolor] (7,3) rectangle +(1,1);
    \fill[freecolor] (8,3) rectangle +(1,1);
    \fill[freecolor] (9,3) rectangle +(1,1);
    \fill[freecolor] (11,3) rectangle +(1,1);
    
    \fill[freecolor] (8,2) rectangle +(1,1);
    \fill[freecolor] (9,2) rectangle +(1,1);
    
    \fill[freecolor] (2,6) rectangle +(1,1);
    \fill[freecolor] (3,6) rectangle +(1,1);
    \fill[freecolor] (4,6) rectangle +(1,1);
    \fill[freecolor] (5,6) rectangle +(1,1);
    \fill[freecolor] (6,6) rectangle +(1,1);
    \fill[freecolor] (7,6) rectangle +(1,1);
    \fill[freecolor] (8,6) rectangle +(1,1);
    \fill[freecolor] (9,6) rectangle +(1,1);
    \fill[freecolor] (10,6) rectangle +(1,1);
    \fill[freecolor] (11,6) rectangle +(1,1);
    
    \fill[freecolor] (5,7) rectangle +(1,1);
    \fill[freecolor] (6,7) rectangle +(1,1);
    \fill[freecolor] (7,7) rectangle +(1,1);
    \fill[freecolor] (8,7) rectangle +(1,1);
    \fill[freecolor] (9,7) rectangle +(1,1);
    \fill[freecolor] (11,7) rectangle +(1,1);
    
    \fill[freecolor] (7,8) rectangle +(1,1);
    \fill[freecolor] (8,8) rectangle +(1,1);
    \fill[freecolor] (9,8) rectangle +(1,1);
    
    \fill[freecolor] (10,9) rectangle +(1,1);
    

    \fill[hitcolor] (11,9) rectangle +(1,1);
    \fill[hitcolor] (12,9) rectangle +(1,1);
    \fill[hitcolor] (13,9) rectangle +(1,1);
    \fill[hitcolor] (14,9) rectangle +(1,1);
    \fill[hitcolor] (15,9) rectangle +(1,1);
    \fill[hitcolor] (10,8) rectangle +(1,1);
    \fill[hitcolor] (11,8) rectangle +(1,1);
    \fill[hitcolor] (12,8) rectangle +(1,1);
    \fill[hitcolor] (13,8) rectangle +(1,1);
    \fill[hitcolor] (14,8) rectangle +(1,1);
    \fill[hitcolor] (15,8) rectangle +(1,1);
    \fill[hitcolor] (10,7) rectangle +(1,1);
    \fill[hitcolor] (11,7) rectangle +(1,1);
    \fill[hitcolor] (12,7) rectangle +(1,1);
    \fill[hitcolor] (13,7) rectangle +(1,1);
    \fill[hitcolor] (14,7) rectangle +(1,1);
    \fill[hitcolor] (15,7) rectangle +(1,1);
    \fill[hitcolor] (12,6) rectangle +(1,1);
    \fill[hitcolor] (13,6) rectangle +(1,1);
    \fill[hitcolor] (14,6) rectangle +(1,1);
    \fill[hitcolor] (15,6) rectangle +(1,1);
    \fill[hitcolor] (12,5) rectangle +(1,1);
    \fill[hitcolor] (13,5) rectangle +(1,1);
    \fill[hitcolor] (14,5) rectangle +(1,1);
    \fill[hitcolor] (15,5) rectangle +(1,1);
    \fill[hitcolor] (12,4) rectangle +(1,1);
    \fill[hitcolor] (13,4) rectangle +(1,1);
    \fill[hitcolor] (14,4) rectangle +(1,1);
    \fill[hitcolor] (15,4) rectangle +(1,1);
    \fill[hitcolor] (10,3) rectangle +(1,1);
    \fill[hitcolor] (11,3) rectangle +(1,1);
    \fill[hitcolor] (12,3) rectangle +(1,1);
    \fill[hitcolor] (13,3) rectangle +(1,1);
    \fill[hitcolor] (14,3) rectangle +(1,1);
    \fill[hitcolor] (15,3) rectangle +(1,1);
    \fill[hitcolor] (10,2) rectangle +(1,1);
    \fill[hitcolor] (11,2) rectangle +(1,1);
    \fill[hitcolor] (12,2) rectangle +(1,1);
    \fill[hitcolor] (13,2) rectangle +(1,1);
    \fill[hitcolor] (14,2) rectangle +(1,1);
    \fill[hitcolor] (15,2) rectangle +(1,1);
    
    \draw[step=4cm,gray,ultra thick] (0,2) grid (16,10);
    \draw[step=2cm,gray,thick] (0,2) grid (16,10);
    \draw[step=1cm,gray,thin] (0,2) grid (16,10);
    
    \fill[deffreecolor] (8,5) rectangle +(1,1);
    \fill[deffreecolor] (9,5) rectangle +(1,1);
    \fill[deffreecolor] (10,5) rectangle +(1,1);
    \fill[deffreecolor] (11,5) rectangle +(1,1);

    \fill[deffreecolor] (9,6) rectangle +(1,1);
    \fill[deffreecolor] (10,6) rectangle +(1,1);
    \fill[deffreecolor] (11,6) rectangle +(1,1);
    
    \fill[deffreecolor] (10,4) rectangle +(1,1);
    \fill[deffreecolor] (11,4) rectangle +(1,1);
    
    
    \path[draw, thick] (sensor) -- ++(0.45,0.3) -- ++(0, -0.6) -- cycle;
\end{tikzpicture}%
        }%
        \caption{}
        \label{fig:method_9}
    \end{subfigure}%
    \caption{Example of integration with larger localization errors (shown as a single slice of the 3D grid). (a) Point cloud with the real sensor position (green) offset one cell up compared to \cref{fig:step_1}, showing that some cells are now incorrectly classified (x). (b) By increasing the number of neighboring cells that must be intersected or hit to two (i.e., \( d_p=2 \)) we account for a localization error of up to two voxels in any direction. This leads to a more conservative classification of void regions (red) compared to \cref{fig:step_2}. (c) Extending the hits away from the sensor to allow classification of void regions next to obstacles (shown in (d)).}
    \label{fig:integration_comparision}
\end{figure}

In the real world, sensor noise and localization errors become a problem. First, we consider the problem related to localization accuracy. 
As mentioned and as in other works, we assume that the sensor pose is given. If the sensor pose is offset from the true pose, the hits and intersected voxels would also be offset, and, by extension, the set of classified void voxels risk being incorrect. An illustration of this problem is shown in~\cref{fig:method_4}, where the true pose is one voxel above the estimated pose. The voxels marked with a cross correspond to voxels that would have changed state given the true pose.

To alleviate this problem, we propose to look not only at the direct neighbors of a voxel, but also at the surrounding voxels at a Chebyshev distance of \( d_p \) away; where \( d_p \) is proportional to the localization error. Setting \( d_p=2 \) in the example shown in~\cref{fig:step_1}, results in the void classified voxels seen in~\cref{fig:method_7}.

From~\cref{fig:method_7}, it is observed that it is now impossible to classify voxels next to hits as void. To deal with this, anything after a hit is also considered a hit (see~\cref{fig:method_8}). This is implemented by extending the ray casting by inserting hits from where the original ray casting ended. The voxels classified as void after this can be seen in~\cref{fig:method_9}.

Lastly, we consider the problem of sensor noise. We modeled sensor noise by marking voxels at a distance \( d_s \) ahead of the hit along the ray as hits. The value of \( d_s \) could depend on the uncertainty of the sensor range per point. In our work, we used a fixed value of \( d_s \).

\subsection{Classifying Points as Dynamic or Static}
The majority of the computations in DUFOMap are associated with the classification of void regions, which is performed once for each new point cloud when it arrives. In contrast, classifying points as static or dynamic requires only a quick lookup of the map. If a point falls into a void voxel (that is, a voxel with \( i_\text{void} = \text{true} \)), it is dynamic, otherwise static. This can be done at any time. By querying as a post-processing step, one benefits from all the information. This is how the map cleaning task is typically done and what we do in the following experiments. In the experiments, we also present DUFOMap\(^{\star}\), which runs online. Here, each new scan is classified using the map that has been built up until that time, whereas DUFOMap can make this decision using a map containing all scans.

\section{Experimental Setup}
We compare our method with the current state-of-the-art represented by Removert~\cite{gskim-2020-iros}, ERASOR~\cite{lim2021erasor}, OctoMap~\cite{hornung13auro} and Dynablox~\cite{schmid2023dynablox}. The first three are evaluated in post-processing mode against DUFOMap. Dynablox is an online method and is compared to DUFOMap\(^{\star}\).

\subsection{Datasets}
To demonstrate that our method can handle a wide range of scenarios and sensor types, we go beyond the use of a dataset with a single sensor type.
To achieve this, we follow the benchmark evaluation protocol presented in \cite{zhang2023dynamic}. The benchmark includes the KITTI dataset~\cite{Geiger2013IJRR} with annotation labels and poses from SemanticKITTI~\cite{behley2019iccv}. We present results on sequences 00 and 01 in the paper. These comprise a small town and a highway, respectively, and are captured with a HDL-64E LiDAR. For other KITTI sequences, we refer to our project page \href{https://kth-rpl.github.io/dufomap}{https://kth-rpl.github.io/dufomap}.
Another dataset~\cite{Argoverse2} collected by two VLP-32C LiDAR sensors is `Argoverse 2 big city' includes various dynamic objects in an urban environment. 
The most sparse one is collected by a 16-channel LiDAR in a highly structured (`Semi-indoor') environment as illustrated in~\cref{fig:full_semindoor}. 

An additional four datasets are used in the qualitative analysis of DUFOMap. We use part of the MCD VIRAL~\cite{mcdviral2024} dataset. It is captured by a Leica RTC360 3D laser scanner commonly used in surveying, where the removal of dynamic objects is also highly relevant.  
It showcases DUFOMap's ability to handle data captured at discrete locations with significant height differences and relatively far apart rather than in a continuous stream as when driving. The data is also very dense (\num{1.3}~million points per scan vs \num{0.1}~million for the 64-channel LiDAR) and the sensor has a much larger vertical field of view (\qty{300}{\degree} vs. \qty{30}{\degree}). 

To showcase the robustness and generalization of our method, we demonstrate our removal performance under highly dynamic and complex environments in~\cref{sec:complex} using datasets collected by a 128-channel LiDAR in a train station and Livox Mid-360~\cite{mid360} in a two-floor structure.
Additionally, small-scale experiments are also presented on our project page where we introduce two datasets captured using a RGB-D sensor.

\subsection{Metric}
Our evaluation metrics from~\cite{zhang2023dynamic} include static accuracy (SA~\%), dynamic accuracy (DA~\%), and associated accuracy (AA~\%) at a point level without downsampling the ground truth map to have an accurate and fair evaluation.
SA represents the proportion of correctly labeled static points, while DA represents the proportion of correctly labeled dynamic points.
\( AA = \sqrt{SA \times DA} \) gives an overall assessment of the algorithm's performance, sensitive to doing well on both SA and DA.

\subsection{Parameter Settings}
To demonstrate our ability to handle different scenarios and sensors in an online setting, where tuning is not possible, we used the same parameter settings for DUFOMap in all experiments where otherwise not stated: voxel size \qty{0.1}{\metre}, \( d_s=\qty{0.2}{\metre} \) for sensor noise, and \( d_p=1 \) for subvoxel localization errors. The same voxel size (\qty{0.1}{\metre}) was used for the other methods. 

For Removert, ERASOR, and OctoMap, we used the parameters from~\cite{zhang2023dynamic}. Meaning, Removert~\cite{gskim-2020-iros} and ERASOR~\cite{lim2021erasor} use per-dataset optimized parameters. 
For Dynablox, we found that the same parameter values, slightly modified from the authors' suggestions, worked well for all experiments. 

\subsection{Hardware}
\label{sec:hw}
Experiments were carried out on a desktop equipped with an Intel Core i9-12900KF. To assess real-time robot applicability, we also performed experiments on a robot equipped with an Intel NUC with an Intel Core i7-8559U.

\begin{table*}[ht!]
\caption{Quantitative comparison of dynamic points removal in point cloud maps. DUFOMap\(^{\star}\), results where we query for each new scan online, using the information acquired so far. The best results are shown in \textbf{bold} and the second best results are shown in \underline{underlined}. Results are in percentage.}
\def\arraystretch{1.3}
\centering
\begin{tabular}{l|rrr|rrr|rrr|rrr} 
\toprule
               & \multicolumn{3}{c|}{KITTI small town (00)}                                                                          & \multicolumn{3}{c|}{KITTI highway (01)}                                                                             & \multicolumn{3}{c|}{Argoverse 2 big city}                                                                                  & \multicolumn{3}{c}{Semi-indoor}                                                                                     \\ 
\hline
Methods        & SA ↑ & DA ↑ & {\cellcolor[rgb]{0.949,0.949,0.949}}AA ↑ & SA ↑ & DA ↑ & {\cellcolor[rgb]{0.949,0.949,0.949}}AA ↑ & SA ↑ & DA ↑ & {\cellcolor[rgb]{0.949,0.949,0.949}}AA ↑ & SA ↑ & DA ↑ & {\cellcolor[rgb]{0.949,0.949,0.949}}AA ↑  \\ 
\hline
Removert~\cite{gskim-2020-iros}   & 99.44                    & 41.53                    & {\cellcolor[rgb]{0.949,0.949,0.949}}64.26                     & 97.81                    & 39.56                    & {\cellcolor[rgb]{0.949,0.949,0.949}}62.20                     & 98.97                    & 31.16                    & {\cellcolor[rgb]{0.949,0.949,0.949}}55.53                     & 99.96                    & 12.15                    & {\cellcolor[rgb]{0.949,0.949,0.949}}34.85                     \\
ERASOR~\cite{lim2021erasor}     & 66.70                    & 98.54                    & {\cellcolor[rgb]{0.949,0.949,0.949}}81.07                     & 98.12                    & 90.94                    & {\cellcolor[rgb]{0.949,0.949,0.949}}\underline{94.46}                     & 77.51                    & 99.18                    & {\cellcolor[rgb]{0.949,0.949,0.949}}87.68                     & 94.90                    & 66.26                    & {\cellcolor[rgb]{0.949,0.949,0.949}}79.30                     \\ 
OctoMap~\cite{hornung13auro}        & 68.05                    & 99.69                    & {\cellcolor[rgb]{0.949,0.949,0.949}}82.37                     & 55.55                    & 99.59                    & {\cellcolor[rgb]{0.949,0.949,0.949}}74.38                     & 69.04                    & 97.50                    & {\cellcolor[rgb]{0.949,0.949,0.949}}82.04                     & 88.97                    & 82.18                    & {\cellcolor[rgb]{0.949,0.949,0.949}}\underline{85.51}                     \\
DUFOMap (Ours) & 97.96                    & 98.72                    & {\cellcolor[rgb]{0.949,0.949,0.949}}\textbf{98.34}            & 98.09                    & 94.20                    & {\cellcolor[rgb]{0.949,0.949,0.949}}\textbf{96.12}            & 96.67                    & 88.90                    & {\cellcolor[rgb]{0.949,0.949,0.949}}\underline{92.70}            & 99.64                    & 83.00                    & {\cellcolor[rgb]{0.949,0.949,0.949}}\textbf{90.94}                     \\
\hline
Dynablox~\cite{schmid2023dynablox}       & 96.76                    & 90.68                    & {\cellcolor[rgb]{0.949,0.949,0.949}}93.67                     & 96.33                    & 68.01                    & {\cellcolor[rgb]{0.949,0.949,0.949}}80.94                     & 96.08                    & 92.87                    & {\cellcolor[rgb]{0.949,0.949,0.949}}\textbf{94.46}                     & 98.81                    & 36.49                    & {\cellcolor[rgb]{0.949,0.949,0.949}}60.05                     \\
DUFOMap\(^{\star}\) (Ours) & 98.37 & 92.37 & {\cellcolor[rgb]{0.949,0.949,0.949}}\underline{95.31} & 98.48 & 81.34 & {\cellcolor[rgb]{0.949,0.949,0.949}}89.50 & 98.66 & 73.98 & {\cellcolor[rgb]{0.949,0.949,0.949}}85.43 & 99.94 & 54.76 & {\cellcolor[rgb]{0.949,0.949,0.949}}73.98
\\
\bottomrule
\end{tabular}
\label{tab:big_num_table}
\end{table*}
\section{Experiments}
\subsection{Quantitative Evaluation}
\subsubsection{Accuracy}
In \cref{tab:big_num_table}, we see that Removert does well in classifying static points (high SA) but struggles to find dynamic points in all datasets (DA \qtyrange[range-units = single, range-phrase=~--~]{20}{40}{\percent}). 
In contrast, both ERASOR and OctoMap detect dynamic points much better, but at the cost of somewhat lower static accuracy, potentially leading to the loss of crucial map features. Our proposed method, DUFOMap, gets high scores on both SA and DA by accurately detecting dynamic points. This enables the generation of complete and clean maps for downstream tasks. The exception is Argoverse 2 where DUFOMap is only the second best for AA and trades the highest value of SA for a slightly lower DA. 

Looking at the online methods, we see that, as expected, DUFOMap\(^{\star}\) performs worse than DUFOMap, which has access to all data before classification is performed. Also, as expected, the performance drop is the smallest in the dataset with the least complex dynamics (KITTI small town). 
Dynablox does the best on the Argoverse 2 dataset where dynamic objects are constantly moving. 

\begin{table}[t]
\caption{Runtime comparison of different methods.}
\centering
\def\arraystretch{1.3}
\begin{tabular}{lcc} 
\toprule
\multicolumn{1}{c}{\multirow{2}{*}{Methods}} & \multicolumn{2}{c}{Run time per point cloud [\unit{\second}] ↓}        \\ 
\cline{2-3}
\multicolumn{1}{c}{}                         & KITTI highway              & Semi-indoor                 \\ 
\hline
Removert~\cite{gskim-2020-iros}   & 0.134 $\pm$ 0.004          & 0.515 $\pm$ 0.024 \\
ERASOR~\cite{lim2021erasor} & 0.718 $\pm$ 0.039          & 0.064 $\pm$ 0.011 \\
OctoMap~\cite{hornung13auro}        & 2.981 $\pm$ 0.952          & 1.048 $\pm$ 0.256 \\
Dynablox~\cite{schmid2023dynablox}       & 0.141 $\pm$ 0.022         & 0.046 $\pm$ 0.008 \\
DUFOMap (Ours) & \textbf{0.062 $\pm$ 0.014}        & \textbf{0.019 $\pm$ 0.003} \\  
\bottomrule
\end{tabular}
\label{time_table}
\end{table}

Both online methods have low DA for the semi-indoor dataset. One of the two people is standing still at the beginning of the dataset and both are still later (see \cref{fig:full_semindoor}(d)). DUFOMap\(^{\star}\) sees these points as dynamic as soon as the person moves for the first time.
Dynablox requires the person to move within a time window to detect it as dynamic. 

Overall, we find that most methods have comparably lower results on the semi-indoor dataset with sparse LiDAR data. 

Across all scenarios and sensor types in various datasets, DUFOMap consistently outperformed the other methods, achieving the highest AA scores and the highest or similar SA and DA scores as the best.

\begin{table*}[h]
\centering
\def\arraystretch{1.3}
\caption{Quantatitive analysis of the influence of pose estimation by different SLAM algorithms for dynamic object removal on KITTI Sequence 00.}
\begin{tabular}{l|ccc|ccc|ccc} 
    \toprule
    & \multicolumn{3}{c|}{GT poses of KITTI \cite{Geiger2013IJRR}}                                                                                           & \multicolumn{3}{c|}{SuMa \cite{behley2019iccv, behley2018rss}}                                                                                  & \multicolumn{3}{c}{KISS-ICP \cite{vizzo2023ral}}                                                                                       \\ 
    \hline
    Methods        & \multicolumn{1}{l}{SA ↑} & \multicolumn{1}{l}{DA ↑} & \multicolumn{1}{l|}{{\cellcolor[rgb]{0.949,0.949,0.949}}AA ↑} & \multicolumn{1}{l}{SA ↑} & \multicolumn{1}{l}{DA ↑} & \multicolumn{1}{l|}{{\cellcolor[rgb]{0.949,0.949,0.949}}AA ↑} & \multicolumn{1}{l}{SA ↑} & \multicolumn{1}{l}{DA ↑} & \multicolumn{1}{l}{{\cellcolor[rgb]{0.949,0.949,0.949}}AA ↑}  \\ 
    \hline
    Removert \cite{gskim-2020-iros}       & 99.18                    & 41.71                    & {\cellcolor[rgb]{0.949,0.949,0.949}}64.32                & 99.44                    & 41.53                    & {\cellcolor[rgb]{0.949,0.949,0.949}}64.26                    & 99.55                    & 41.45                    & {\cellcolor[rgb]{0.949,0.949,0.949}}64.23                                \\
    ERASOR \cite{lim2021erasor}        & 63.83                    & 98.35                    & {\cellcolor[rgb]{0.949,0.949,0.949}}79.23                & 66.70                    & 98.54                    & {\cellcolor[rgb]{0.949,0.949,0.949}}81.07                    & 67.86                    & 98.68                    & {\cellcolor[rgb]{0.949,0.949,0.949}}81.83                                \\
    Octomap \cite{hornung13auro}       & 54.81                    & 99.56                    & {\cellcolor[rgb]{0.949,0.949,0.949}}73.87                & 68.05                    & 99.69                    & {\cellcolor[rgb]{0.949,0.949,0.949}}82.37                    & 62.28                    & 99.85                    & {\cellcolor[rgb]{0.949,0.949,0.949}}78.85                                \\
    Dynablox \cite{schmid2023dynablox}      & 95.50                    & 89.34                    & {\cellcolor[rgb]{0.949,0.949,0.949}}\underline{92.37}                & 96.76                    & 90.68                    & {\cellcolor[rgb]{0.949,0.949,0.949}}\underline{93.67}                    & 98.31                    & 90.97                    & {\cellcolor[rgb]{0.949,0.949,0.949}}\underline{94.57}                                \\
    DUFOMap (Ours) & 92.57                    & 98.52                    & {\cellcolor[rgb]{0.949,0.949,0.949}}\textbf{95.50}       & 97.96                    & 98.72                    & {\cellcolor[rgb]{0.949,0.949,0.949}}\textbf{98.34}           & 99.33                    & 98.73                    & {\cellcolor[rgb]{0.949,0.949,0.949}}\textbf{99.03}                       \\
    \bottomrule
    \end{tabular}
    \label{tab:pose}
\end{table*}

\subsubsection{Execution Time}
\Cref{time_table} presents information on the run time of the different methods for two of the datasets, one with a 64-channel LiDAR (KITTI highway) and one with a 16-channel LiDAR (semi-indoor). The reported time is the total processing time divided by the number of point clouds. Note that this is mainly important for the three latter methods, for which an online mode is supported. However, we see that OctoMap is prohibitively slow, requiring \qty{3}{\second} to integrate a single frame in KITTI. 

In the case of the sparse LiDAR (semi-indoor), most methods operate faster due to the reduced number of points in a single scan (64 $\rightarrow$ 16) and the shorter sensor range.
In general, our method outperforms other methods in both dense and sparse sensor settings.

Inspired by Dynablox~\cite{schmid2023dynablox}, we also performed a test on a low-power computer commonly found on robots (an Intel® NUC, see~\cref{sec:hw}). We used the Dynablox setup and reduced the range to \qty{20}{\metre}. DUFOMap maintained a frequency of \qty{20}{\hertz} on the 4-core CPU on the semi-indoor dataset. In comparison, Dynablox operated at less than \qty{10}{\hertz}.

\subsection{Qualitative Results}
In this section, we analyze the performance of our method on additional datasets to demonstrate that our method is capable of handling different sensors and scenarios. Note that we use the same setting for the parameters as in previous experiments.

\begin{figure*}[t]
  \centering
  \includegraphics[trim=1 60 1 15, clip, width=\linewidth]{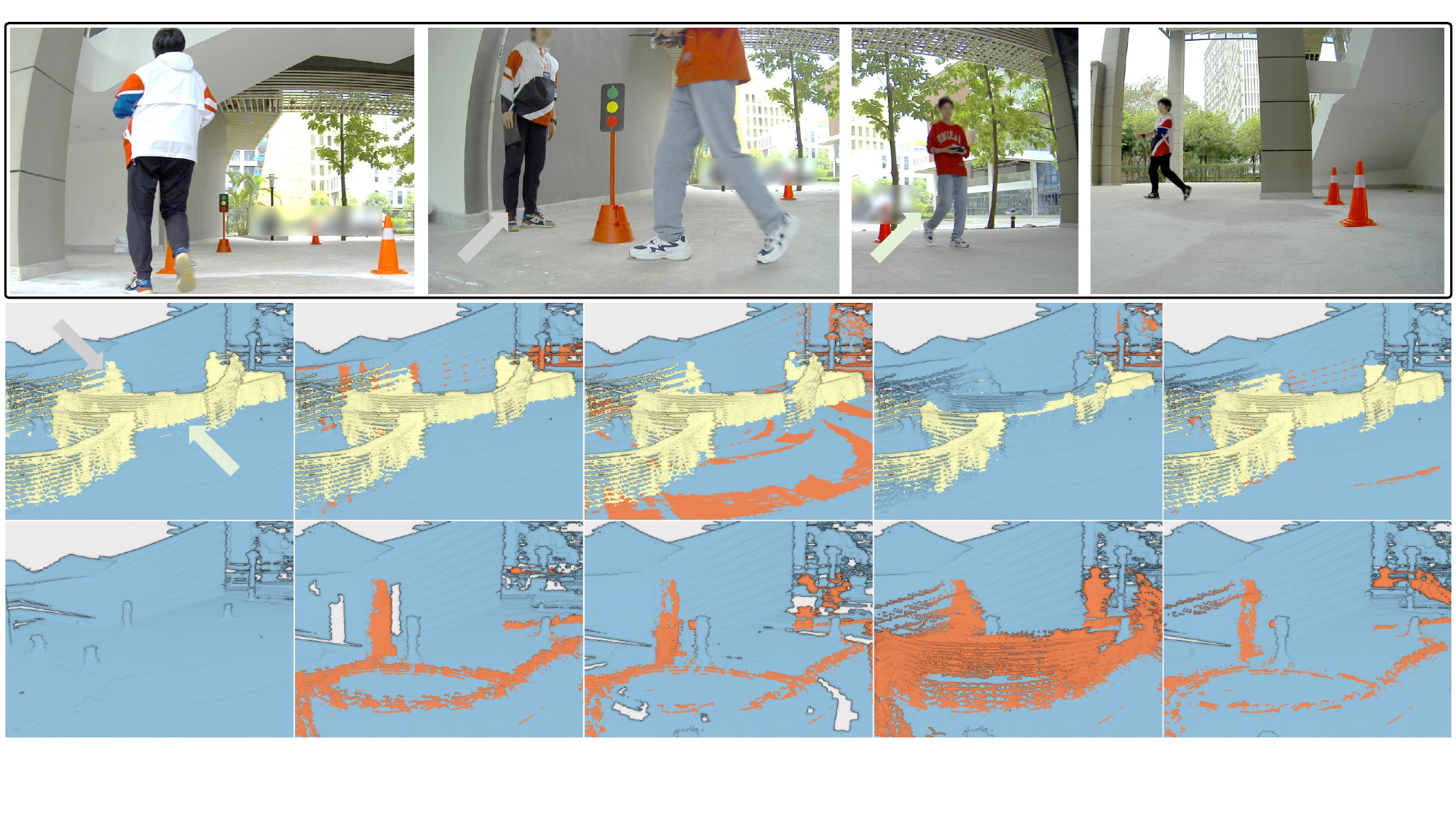}
  \newcolumntype{C}{>{\centering\arraybackslash}X}
\begin{tabularx}{\linewidth}{@{} *{5}{C} @{}}
  \small (a) Ground Truth & \small (b) ERASOR~\cite{lim2021erasor} & \small (c) OctoMap~\cite{hornung13auro} & \small \hspace{-15pt}(d) Dyanblox~\cite{schmid2023dynablox} & \small \hspace{-5pt}(e) 
  DUFOMap (Ours) \\[6pt]
\end{tabularx}
\vspace{-10pt}
\caption{\small 
    Qualitative results from a self-collected dataset with a sparse LiDAR sensor (VLP-16). 
    Points labeled as true positives are colored yellow, whereas incorrectly classified points are colored orange. 
    The first column provides ground truth labels obtained by human annotation. The third row presents the clean map output from different methods, with orange marking any remaining dynamic points in the output map.}
  \label{fig:full_semindoor}
\end{figure*}

In the semi-indoor environment (\cref{fig:full_semindoor}), two people move around the sensor.
It was harder to find good height thresholds for ERASOR in this scenario. We can see some points belonging to heads and feet still remaining. OctoMap has problems with the sparse LiDAR data (16 channels), where rays occasionally penetrate the ground, and thus to erroneous removal. The ground in their output map displays LiDAR ring-like gaps. 
Dynablox's performance on this dataset is markedly worse. Especially the dynamic accuracy is low, which is clearly visible as a few yellow points in the second row (few true positives) and many orange points in the third (many false negatives).
Like in the previous scenario, DUFOMap gives the best output with high accuracy in both dynamic and static parts. 
Accounting for sensor noise and some localization errors allows DUFOMap to handle the issues faced by OctoMap regarding the ground points and to accurately identify void regions and thus dynamic points.

\subsubsection{Highly Dynamic and Complex Environments}
\begin{figure}
    \centering
    \begin{subfigure}[b]{\linewidth}
        \includegraphics[width=\textwidth]{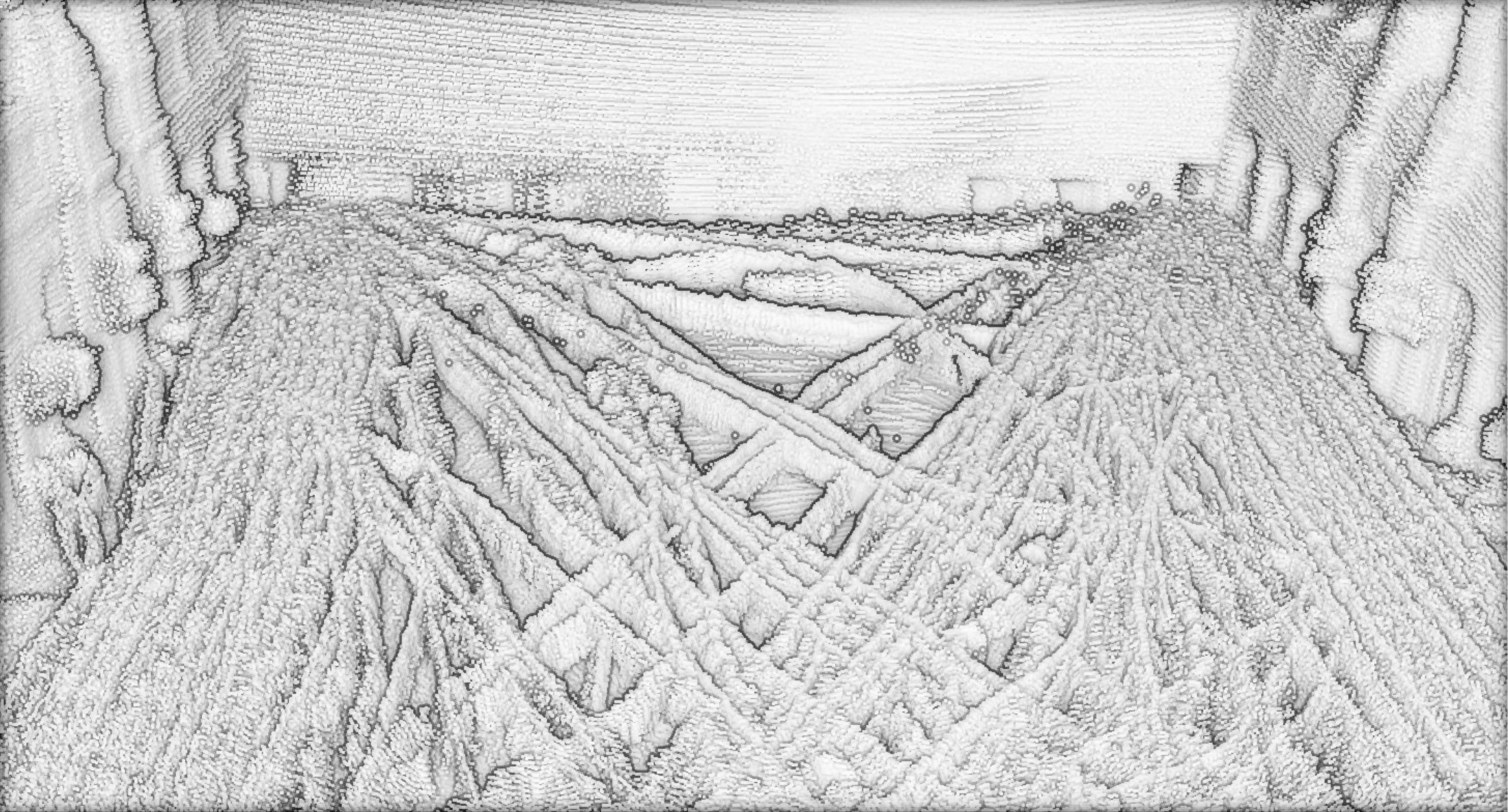}
        \caption{Raw (Unclean) Map}
        \label{fig:raw_doals}
    \end{subfigure}
    \hfill
    \begin{subfigure}[b]{\linewidth}
        \includegraphics[width=\textwidth]{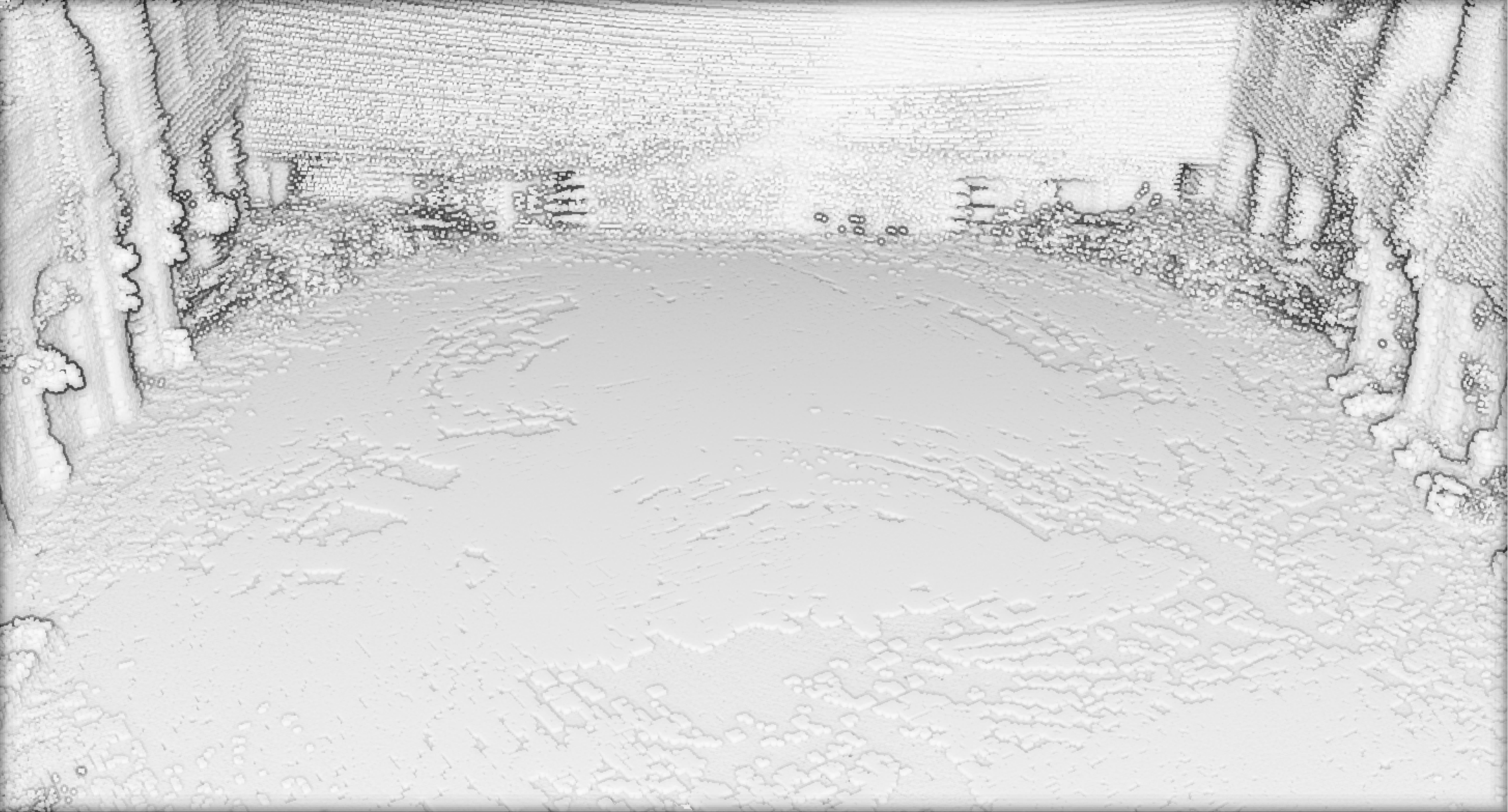}
        \caption{Cleaned Map}
        \label{fig:dufo_doals}
    \end{subfigure}
    \caption{DUFOMap dynamic points removal performance on DOALS Dataset \cite{pfreundschuh2021dynamic} (highly dynamic environment).}
    \label{fig:doals_demo}
\end{figure}

In~\cref{fig:doals_demo}, results are shown from the Urban Dynamic Objects LiDAR (DOALS) Dataset~\cite{pfreundschuh2021dynamic}, which is collected in a highly dynamic train station environment. After DUFOMap processing, we get a nicely cleaned map even in this complex and highly dynamic environment.

\begin{figure}
    \centering
    \color{blue}
    \begin{subfigure}[b]{0.48\linewidth}
        \includegraphics[width=\textwidth]{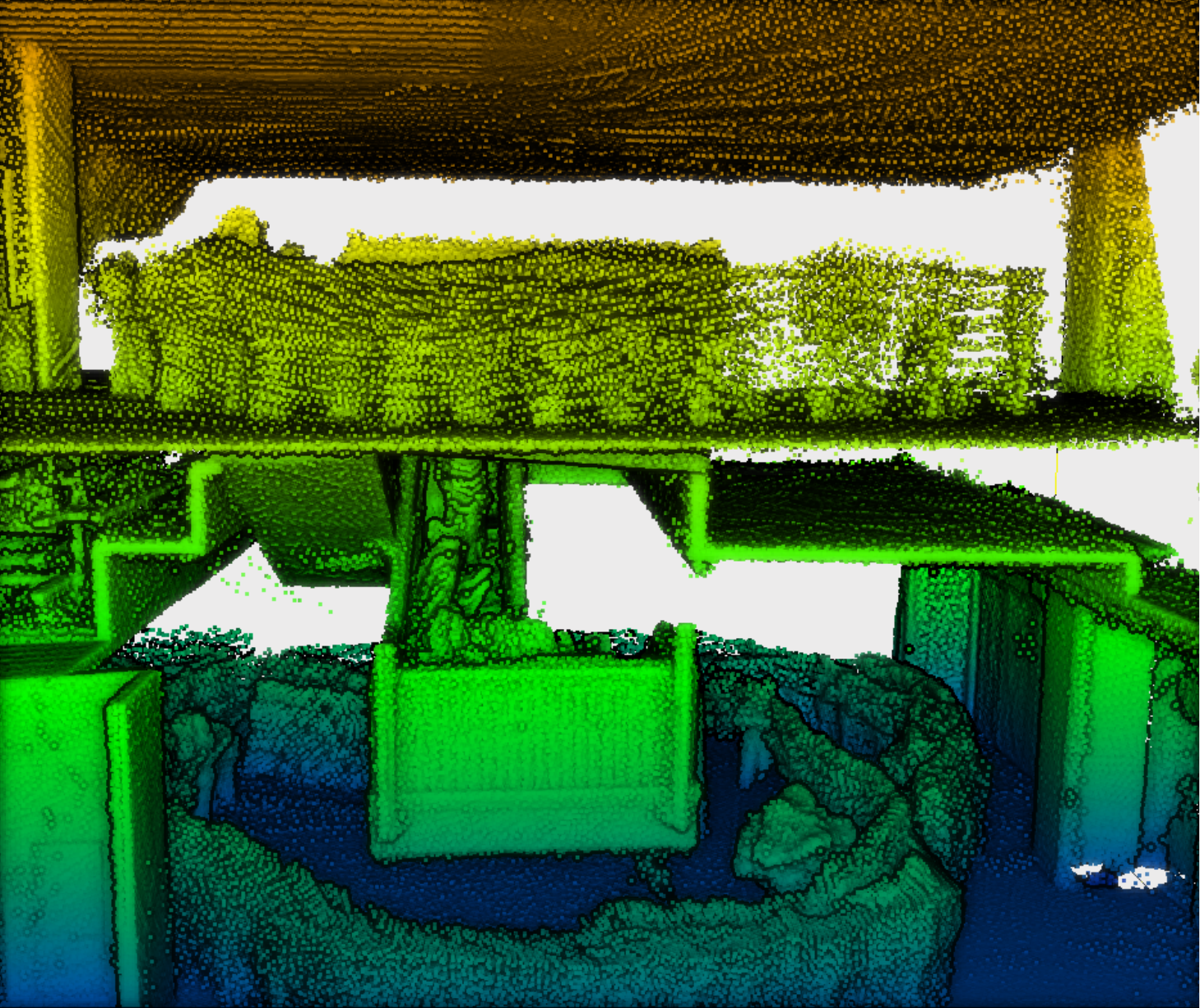}
        \caption{Raw (Unclean) Map}
        \label{fig:raw_stair}
    \end{subfigure}
    \hfill
    \begin{subfigure}[b]{0.48\linewidth}
        \includegraphics[width=\textwidth]{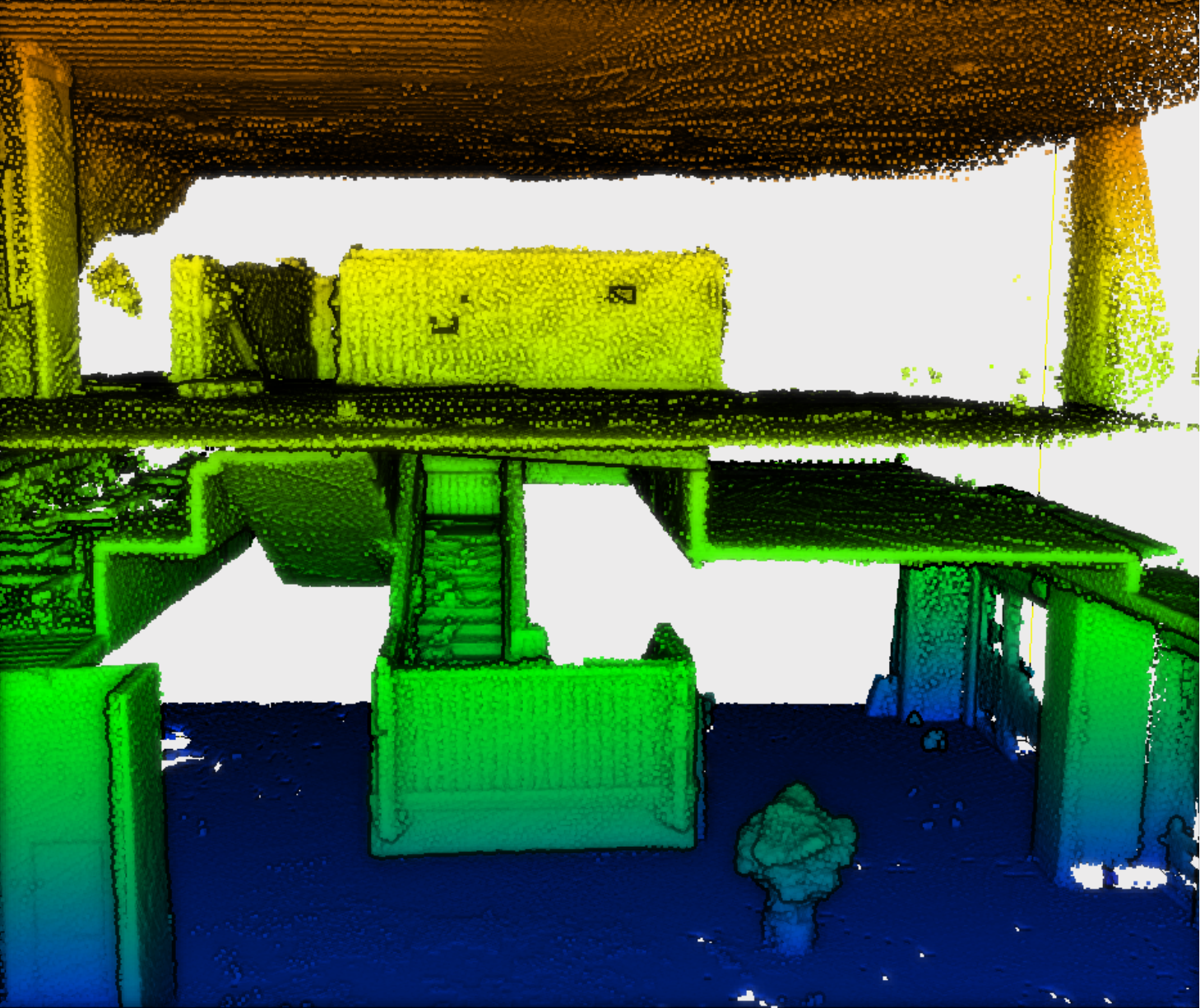}
        \caption{Cleaned Map}
        \label{fig:dufo_stair}
    \end{subfigure}
    \caption{DUFOMap dynamic points removal performance in a complex, two-floor, structure. The color indicates different heights. For clearer visualization, parts of the walls have been removed.}
    \label{fig:stair}
    \vspace{-1em}
\end{figure}

\Cref{fig:stair} shows a scene from a two-floor building that challenges methods that make assumptions about the height, the ground level, etc. DUFOMap is able to effectively remove dynamic points.

\subsubsection{Survey Sensor Dataset}
\label{sec:complex}
Our method is capable of removing the dynamic points from the Leica dataset, seen in~\cref{fig:background}. 
The dataset is difficult for the other methods. The following is an analysis based on the theory of the other methods to unravel these difficulties. The probabilistic model of OctoMap suffers from the few samples per voxel.
For ERASOR, the height threshold would have to be automatically adjusted or redefined to handle the height differences. In a driving scenario on horizontal ground with a fixed sensor height, a fixed threshold can give very good results. However, as the height of both the ground and from which the data is captured changes, as in surveying data, a fixed threshold is no longer possible. Dynablox cannot handle non-sequential data.

The output of our method is depicted in~\cref{fig:background} (lower right) when applied to the raw data (upper). 
In the detection results (lower left), we naively clustered the points so that different objects stand out. 
For a comprehensive view of our results, readers are encouraged to visit our project page\footnote{\href{https://kth-rpl.github.io/dufomap}{https://kth-rpl.github.io/dufomap}} where a detailed video of the entire process is available for viewing.

\subsection{Influence of Pose Estimation}
In this section, we study how the pose influences the results. We investigate three different sources of pose estimates. The KITTI dataset comes with poses that are used as ground truth in the KITTI odometry benchmark. The ground truth poses of the SemanticKITTI dataset are estimated by SuMa~\cite{behley2018rss}. The third method is KISS-ICP~\cite{vizzo2023ral}. In~\cref{tab:pose} we can see that all methods are influenced by the pose. With worse pose estimates, the scenes will appear more dynamic. Therefore, it will be more difficult to classify the static part, and we would expect SA to decrease with increased pose errors. Based on this, KISS-ICP provides more accurate and consistent pose estimates for the short sequences used in the benchmark, closely followed by the ground truth pose of SemanticKITTI (SuMa).

\begin{table}[t]
\centering
\def\arraystretch{1.3}
\caption{Ablation study of DUFOMap. \( v \) voxel size [\unit{\metre}].}
\begin{tabular}{l ccc} 
\toprule
Parameter settings & \multicolumn{1}{l}{SA [\%] ↑} & \multicolumn{1}{l}{DA [\%] ↑} & \multicolumn{1}{l}{{\cellcolor[rgb]{0.949,0.949,0.949}}AA [\%] ↑}  \\ 
\midrule
w/o $d_s$, $d_p$, $v=0.1$     & 14.89                    & 99.99                    & {\cellcolor[rgb]{0.949,0.949,0.949}}38.58                     \\
$d_s=0.2$, $v=0.1$           & 30.29                    & 99.99                    & {\cellcolor[rgb]{0.949,0.949,0.949}}55.03                     \\
$d_p=1$, $v=0.1$           & 91.89                    & 98.97                    & {\cellcolor[rgb]{0.949,0.949,0.949}}95.37                     \\ 
$d_s=0.2$, $d_p=1$, $v=0.2$	& 92.97	& 98.24	& {\cellcolor[rgb]{0.949,0.949,0.949}}95.57 \\
\hline
$d_s=0.2$, $d_p=1$, $v=0.1$ & 97.96                    & 98.72                    & {\cellcolor[rgb]{0.949,0.949,0.949}}\textbf{98.34}                     \\
\bottomrule
\end{tabular}
\label{ablation_table}
\vspace{-1em}
\end{table}

\subsection{Ablation Study}
\label{sec:ablation}
In this section, we will look at how the parameters corresponding to the localization error (\( d_p \)) and the sensor noise (\( d_s \)) influence the behavior.

In our early experiments, we noted that many methods struggle (not surprisingly) to correctly classify the points on the border between what is static and what is dynamic. A person's feet touching the ground is a common and challenging example. In earlier versions of our algorithm, we used clustering to address this problem. We later find that we achieve a more accurate and general solution by better modeling the data, capturing localization errors with \( d_p \) and sensor noise with \( d_s \). 
In \cref{ablation_table} we present the results from an ablation study on the KITTI sequence 00 where we turn on and off the sensor noise and localization error models controlled by the \( d_p \) and \( d_s \) parameters. The results clearly show how important it is to correctly classify the void regions, which is the foundation of our method. When we do not model the errors, more points will be incorrectly classified as dynamic (DA increases). 
In the first row, we see an extreme example where the static map is almost empty (very low SA). We can also see that accounting for localization errors (with \( d_p \)) has the greatest impact. This is because it affects the classification along the whole ray, whereas the sensor noise model mainly affects regions close to the hit.

\subsection{Limitations}
While we can operate on a large number of scenarios and sensors with the same parameter settings, there are limitations. When the LiDAR data is sparse, our conservative model for classifying void regions leads to lower dynamic accuracy. We require a certain density to ensure that neighboring voxels are observed. This is seen in both the semi-indoor dataset (16-channel) and Argoverse 2 (32-channel but very long distances).
Based on our void space classification strategy, our method must see at least the region as void once to determine that the points inside the region are dynamic. This may lead to limitations that, for a big bus moving slowly, part of the points may not be classified as dynamic. To remedy the limitations mentioned above, we can involve clustering~\cite{campello2013density} or integrate with learning-based detection~\cite{MMDetection_Contributors_OpenMMLab_Detection_Toolbox_2018} or scene flow estimation~\cite{zhang2024deflow} to improve the whole pipeline in future work.

\section{Conclusion}
In this work, we have presented DUFOMap as a dynamic awareness method based on UFOMap. Dynamics is identified implicitly by classifying empty regions of the environment. DUFOMap was evaluated against four state-of-the-art methods, in multiple different scenarios, and with a variety of sensors, showing the best overall performance.

\section*{Acknowledgment}
Thanks to HKUST Ramlab's members: Bowen Yang, Lu Gan, Mingkai Tang, and Yingbing Chen, who help collect additional datasets. 
We also thank the anonymous reviewers for their constructive comments.

\bibliographystyle{IEEEtran}
\bibliography{IEEEabrv,ref}
\end{document}